\documentclass[letterpaper, 10 pt, conference]{ieeeconf}  

\IEEEoverridecommandlockouts                              

\usepackage{multicol}
\usepackage{xcolor}
\usepackage{booktabs} 
\definecolor{lightblue}{rgb}{0,0.2,1}
\definecolor{black}{rgb}{0,0,0}
\hypersetup{
    colorlinks=true,%
    citebordercolor=white,%
    filebordercolor=white,%
    linkbordercolor=white,%
    urlbordercolor=white,%
    citecolor=lightblue,%
    linkcolor=lightblue,%
    urlcolor=lightblue%
}

\usepackage{amsmath,amssymb,mathrsfs}
\usepackage[noend]{algorithmic}
\usepackage[ruled, linesnumbered, vlined]{algorithm2e}
\usepackage[Symbol]{upgreek}
\usepackage{comment}
\usepackage{graphicx}
\usepackage{graphics}
\usepackage[tight]{subfigure}
\usepackage{float}
\usepackage{multirow}
\usepackage{array}
\usepackage{enumerate}
\usepackage{sidecap}
\usepackage[all]{xy}
\graphicspath{{./}}
\usepackage{changepage}   
\usepackage{wrapfig}
\usepackage{flushend}

\newtheorem{proposition}{\textbf{Proposition}}

\usepackage{array}
\newcolumntype{L}[1]{>{\raggedright\let\newline\\\arraybackslash\hspace{0pt}}m{#1}}
\newcolumntype{C}[1]{>{\centering\let\newline\\\arraybackslash\hspace{0pt}}m{#1}}
\newcolumntype{R}[1]{>{\raggedleft\let\newline\\\arraybackslash\hspace{0pt}}m{#1}}

\usepackage{color}

\newcounter{tecounter}
\title{\LARGE \bf
A Spatio-Temporal Representation for the Orienteering Problem \\ with Time-Varying Profits }

\author{Zhibei Ma, Kai Yin,  Lantao Liu, Gaurav S. Sukhatme
\thanks{Z. Ma, L. Liu and G.S. Sukhatme are with the Department of Computer Science at the University of Southern California, Los Angeles, CA 90089, USA. {\tt\small \{zhibeima, lantao.liu, gaurav\}@usc.edu}. 
K. Yin is with HomeAway, Inc., {\tt\small yinkai1000@gmail.com}.
}
}

\begin{document}

\maketitle
\thispagestyle{empty}
\pagestyle{empty}

\begin{abstract}
We consider an orienteering problem (OP) where an agent needs to visit a series (possibly a subset) of depots, from which the maximal accumulated profits are desired within given limited time budget.
Different from most existing works where the profits are assumed to be static, in this work we investigate a variant that has arbitrary time-dependent profits.
Specifically, the profits to be collected change over time and they follow different (e.g., independent) time-varying functions. 
The problem is of inherent nonlinearity and difficult to solve by existing methods. 
To tackle the challenge, we present a simple and effective framework that incorporates time-variations into the fundamental planning process.
Specifically, we propose a deterministic {\em spatio-temporal} representation where both spatial description and temporal logic are unified into one routing topology.
By employing existing basic sorting and searching algorithms, the routing solutions can be computed in an extremely efficient way. 
The proposed method is easy to implement and extensive numerical results show that our approach is time efficient and generates near-optimal solutions.


\end{abstract}


\section{Introduction}

The rapid progress of smart vehicle technologies allows us to envision that, in the future autonomous vehicles are able to carry out various tasks with little or even no human effort. 
We are interested in designing an efficient routing method to navigate a vehicle (agent) among a number of known and fixed depots, where each depot has some profit (e.g., score, benefit, utility, load) to be collected. 
If the travel time is limited, it is likely that the agent is not able to traverse all depots due to the limited time budget. This variant of routing problems is called the {\em orienteering problem (OP)}~\cite{golden1987orienteering}.
In a nutshell, an OP aims to find a tour traversing a subset of depots so that the accumulated  profit collected from those traversed depots are maximized.

Different from many existing vehicle routing problems which focus on analyzing path properties in the static context (i.e., unchanging environment or topology with stationary cost metrics)~\cite{TSP-survey92,gunawan2016orienteering}, 
in this work we are intrigued to investigate a time-varying variant of OPs, i.e., each depot has a time-varying profit. 

\begin{figure}
  \centering
   	\includegraphics[height=1.2in]{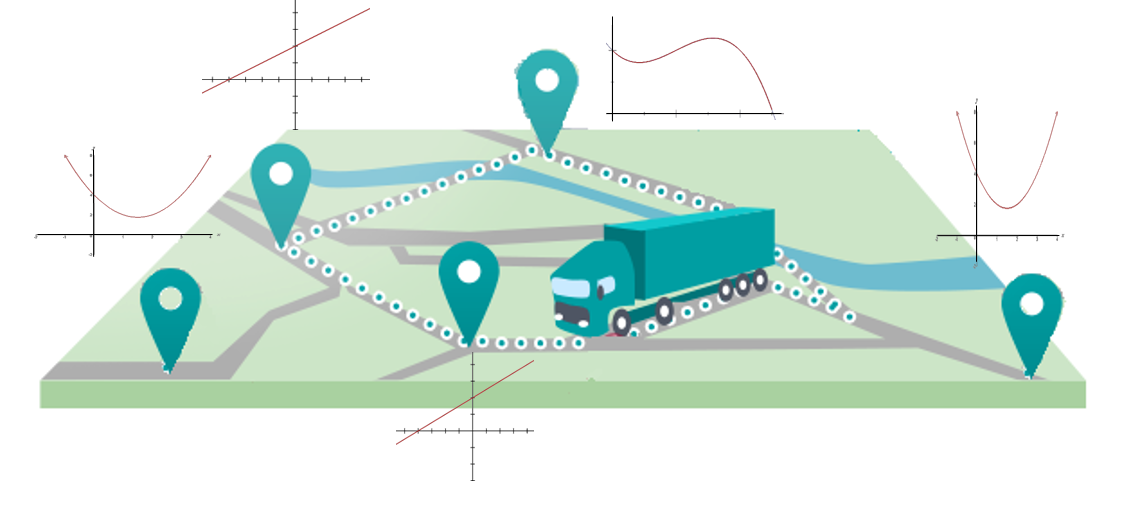}  \vspace{-8pt}
   \caption{An autonomous vehicle needs to route through depots with time-varying profits (loads). The goal is to maximize overall collected profits within given limited time. The task can be persistent.}
\label{fig:example}
\end{figure}

Here are a few motivational examples:  
\begin{itemize}
    \item 
    As illustrated in Fig.~\ref{fig:example}, an autonomous truck needs to pick up goods from a number of fixed depots where  manufacturing factories are located. The goods are produced consistently and accumulated as time goes by --- so they are time-varying. The growth rate of goods at different factories may be  nonidentical: larger factories grow faster and smaller ones grow slower. Assume the truck has sufficiently large capacity and it empties all goods from a depot when it arrives there. The objective is to find a route so that the truck will load the most goods in a given time window.
    \item In environmental monitoring, autonomous robots are deployed to collect environment data in order to estimate an underlying environment state. However, the environment attributes at different locations can be time-varying (e.g., dissolved chemical compounds and algae blooms in the water vary both spatially and temporally.) 
    An important objective of environment monitoring is to plan so-called ``informative paths"~\cite{binney2013optimizing} that navigate the robots to acquire data from those most information-rich spots which best help estimate the environment. 
\end{itemize}

In this paper, we present our initial study for the time-varying OP.  
We start our analysis from the single agent planning case, and attempt to establish a new framework that is different from traditional modeling and solving routines. 
We will show that, a big challenge of the time-varying OP (even for single agent case) lies in the requirement of a special treatment in an extra dimension --- the time dimension, which inevitably introduces extra complexity as we need to model, predict, and integrate future dynamics.


\section{Related Works}

The routing problems have been well studied in many research domains including operations research, theoretical computer science, and transportation systems~\cite{bellman1958routing,TSP-survey68,christofides1976vehicle}. 
Typical routing problems involve incorporation of constraints expressed from the nature of the target problems, which also narrow the space of searching for solutions~\cite{miller1960integer}. 
We are interested in a variant of routing problems called the orienteering problems (OPs)~\cite{golden1987orienteering,vansteenwegen2011orienteering}. 
An OP considers both travel cost (e.g., travel time) and scores collected along the travel.
The goal of the OP is to determine which subset of vertices to visit and in which order so that the collected scores within a given period is maximized. 
The OPs integrate characteristics of both knapsack problems (KPs)~\cite{kellerer2004introduction} and travelling salesman problems (TSPs)~\cite{miller1960integer}, and OPs are NP-hard as well. In contrast to the TSP, not all vertices of an OP need to be visited due to the limited time budget.
During the past few decades, several variants of OPs have been studied, such as time-dependent OPs, Team OPs,  (Team) OPs with Time Windows and OPs with stochastic profits.
Recent survey papers~\cite{vansteenwegen2011orienteering,gunawan2016orienteering} have profoundly discussed state-of-the-art techniques of these variants as well as their applications.

While OPs can be formulated as mixed integer programs (MIPs), the problem size typically is too large to directly use commercial solvers.
A wide range of decomposition methods such as branch-and-price algorithms have been developed so that a large-scale MIP can be decomposed into smaller problems (e.g., a master problem and a series of sub-problems) which can then be iteratively solved by commercial solvers~\cite{gutierrez2010branch, keshtkaran2016enhanced}. 
In order to reduce the heavy computational burden in decomposition methods, heuristics and metaheuristics have been extensively studied, typically including tabu-based or neighborhood search based
procedures~\cite{chao1996team, tang2005tabu,vidal2015large}.

Although in general OPs have been well researched, the time-dependent OPs (TOPs) have received relatively less attention comparing with other variants~\cite{garcia2010hybrid,gunawan2016orienteering}.
Even so, most of existing TOPs discuss the time-varying properties that are associated with the real travel time between pairwise nodes, and assume that travel time between two nodes depends on the departure time at the first (or an earlier) 
node~\cite{mahmoudi2016finding,fomin2002approximation,de2010hybrid,verbeeck2014fast}.
Very rarely we could find the works that discuss about time-varying scores of OPs.
One work that share certain similarity with this proposed problem is~\cite{afsar2013team}, where multiple vehicles need to serve a number of clients and the profit of each client follows a decreasing function of time. 
The work analyzed a lower bound and upper bound based on a classic MIP formulation.

Instead of employing conventional techniques such as the column generation approach used in \cite{afsar2013team}, 
in this work we present our first study that models and tackles the problem from a different perspective: we start from establishing a 
representation built from the spatial and temporal constraints, so that the time dependence attribute is transferred from the bulk MIP to a separate and intuitive representation. 
With that, fast approximate OP solutions can be found by employing and extending order-sensitive topological algorithms.


\section{Problem Description and Formulation}

A routing problem can be represented with a graph $G=(V,E)$, where $V$ is the set of vertices and $E$ denotes the set of edges. Let us denote the number of vertices as $|V| = n+1$, and suppose every edge takes time to traverse.
Although the travel time, in many situations, depends on the states and properties of the network such as congestion and capacity, we assume that the travel time $\tau_{ij}$ between two vertices $v_i$ and $v_j$ is time-invariant for simplicity\footnotemark. 
\footnotetext{It is relatively straightforward to incorporate time-dependent travel times in our proposed model described in the following section.} 
We associate each vertex $v_i$ with a time-varying value, called profit, 
which is denoted by $f_i(t)\geq 0$ at time $t \geq 0$. Here $f_i(\cdot)$ is of arbitrary nonlinear function form. 
We assume that function $f_i(\cdot)$ is known or can be predicted or approximated.

Suppose that the agent (vehicle) starting from a dummy node $v_0$ at time $0$ travels across a subset of vertices on the graph $G$. When the agent visits vertex $v_i$ at time $t_i$, it will collect the profit $f_i(t_i)$. The remaining profit at the vertex $v_i$ right after the agent leaves becomes $0$ and accumulates again. Additionally, the profit at $v_0$ is assumed to be $0$, i.e., $f_0(t)=0$.
Assuming the agent visits each vertex once, the objective is to determine the order of a subset of vertices to visit so that total profits collected by the agent is maximized within a given planning period $T$. 
Note that we do not assign a specific destination to the agent, as the problem with a fixed end vertex is a special case for our problem. 

\begin{figure*}[t] \vspace{-10pt}
  \centering
 \subfigure[]
 	{\label{fig:n3t4_2}\includegraphics[height=1.3in]{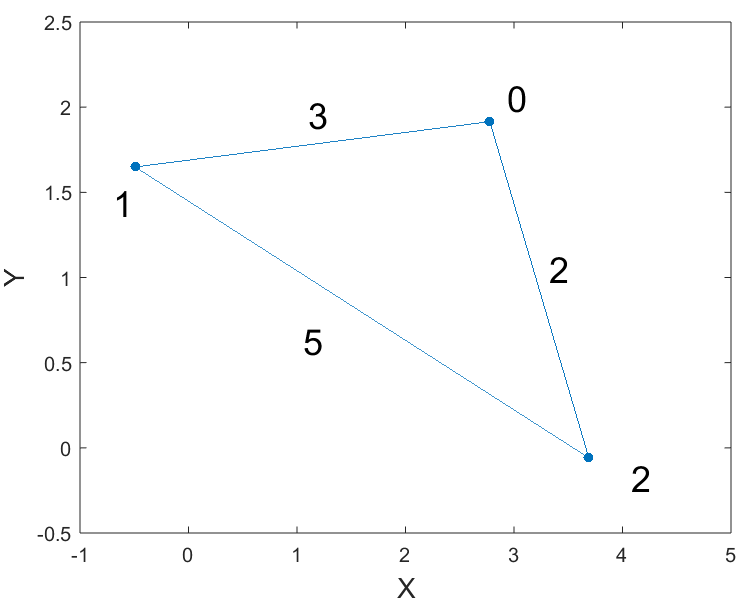}}
 \subfigure[]
 	{\label{fig:n3t4}\includegraphics[height=1.3in]{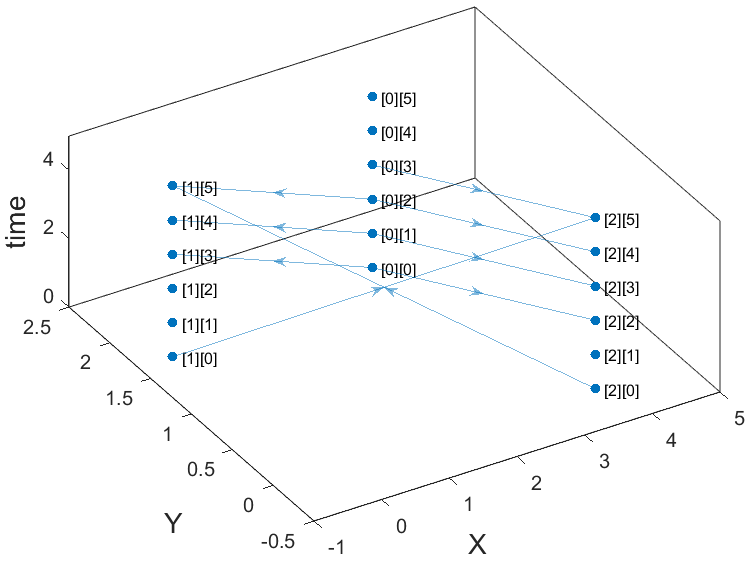}}
  \subfigure[]
 	{\label{fig:n7t7_2}\includegraphics[height=1.3in]{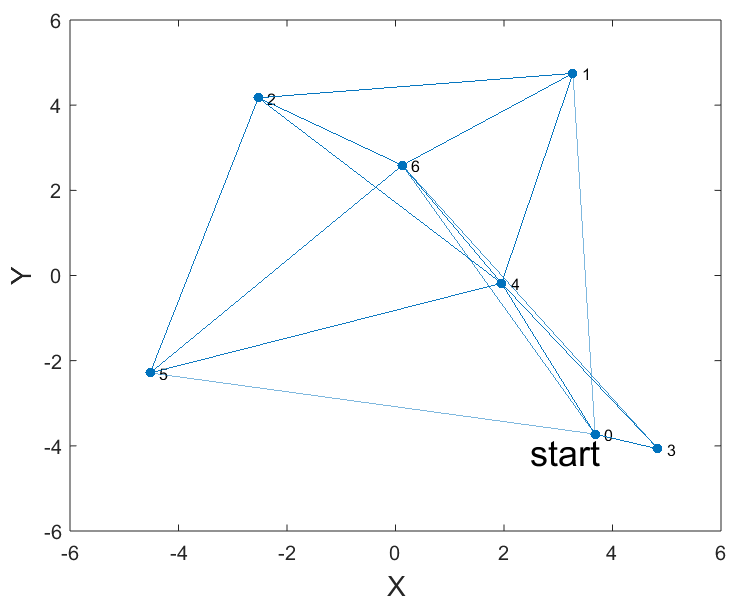}} 
  \subfigure[]
 	{\label{fig:n7t7_3}\includegraphics[height=1.3in]{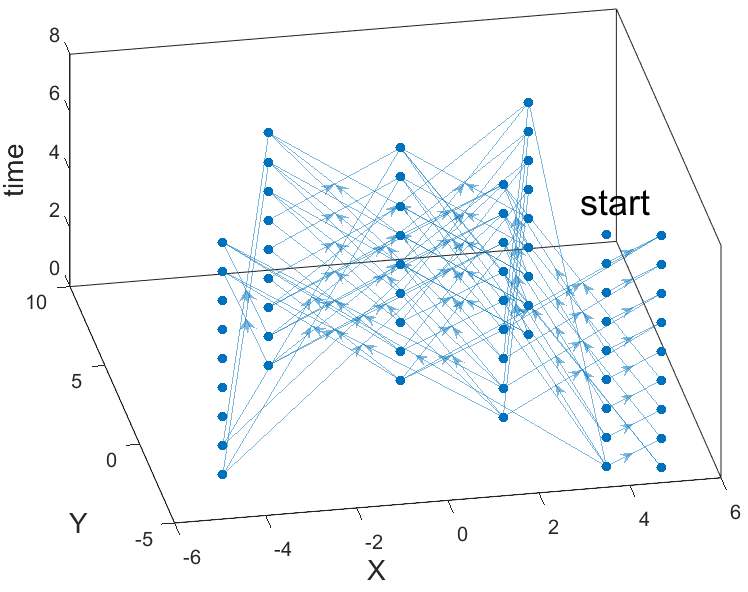}} \vspace{-8pt}
  \caption{Illustrations of spatio-temporal representation. (a)\, Graph $G= (V, E)$ constructed on a $x$-$y$ plane. Numbers on vertices are the vertex IDs, and numbers on edges denote required travel time costs; 
  (b)\, Spatio-temporal graph of (a). In this example each vertex has 5 time units along the ($z$-axis); (c)(d) Another example with 7 vertices and 8 time windows.  
  }
\label{fig:time_model} \vspace{-10pt}
\end{figure*}

Let $x_{ij}=1$ if the agent travels from $v_i$ to $v_j$, and 0 otherwise. 
Let $t_0 = 0$. We also introduce extra variables $u_i$ with $u_0 = 0$ to eliminate the subtours. 
Then the problem in this study can be formulated as the following mixed-integer program.
\begin{align}
    \max_{x_{ij}} &\sum\limits_{i=1}^{n} f_{i}(t_i)\cdot\sum_{j}x_{ji}, \label{eqt:f5}\\
    &\sum\limits_{j=1}^{n} x_{0j} =\sum\limits_{i=1}^{n} x_{i0} = 1, \label{eqt:f6}\\
    &\sum\limits_{i=0}^{n} x_{ik} = \sum\limits_{j=0}^{n} x_{kj} \leqslant 1, \quad \forall k\neq 0, \label{eqt:f7}\\
    &t_j\leqslant T\cdot\sum\limits_{i=0}^{n}x_{ij}, \quad \forall j, \label{eqt:f8}\\
    &t_i + x_{ij}\left(\tau_{ij} + T\right) \leqslant T + t_j, \quad \forall i,j, \label{eqt:f8b}\\
    &1 \leqslant u_i \leqslant n, \quad \forall i\neq 0, \label{eqt:f9}\\
    &u_i-u_j+1 \leqslant n(1-x_{ij}), \quad \forall i\neq 0,j\neq 0, \label{eqt:f10}\\
    &x_{ij} \in \{0,1\}, \quad \forall i,j. \label{eqt:f11}
\end{align}
Eq.~\eqref{eqt:f5} is to maximize the total collected profits. Constraint~\eqref{eqt:f6} guarantees that the path starts and ends at dummy node $v_0$. Note that our setup allows a path 
 to end at any node by discarding the inward edge. 
Constraints~\eqref{eqt:f7}
ensure the connectivity of the path and guarantee that every vertex is visited at most once.
The limited time budget is ensured by Constraint~\eqref{eqt:f8}, while Constraint~\eqref{eqt:f8b} determines the visiting time $t_i$ along the path. Constraints~\eqref{eqt:f9} and~\eqref{eqt:f10}, so called Miller-Tucker-Zemlin (MTZ) formulation in the traveling sales problem\cite{miller1960integer}, are used to prevent subtours.


\section{A Spatio-Temporal Representation}

It appears a daunting task to solve the problem ~\eqref{eqt:f5}-\eqref{eqt:f11} due to the complexity of objective function. Most of existing literature in OPs either lumps all constraints together and solves it by conventional solvers of MIPs~\cite{nemhauser1988integer, vansteenwegen2011orienteering}, or decouples the constraints into master-subproblem modules such as column generation~\cite{afsar2013team}, or use certain heuristics such as center-of-gravity heuristic ~\cite{golden1987orienteering}. 
In contrast, we desire to develop a framework starting from a constraint-included representation that is intuitive to understand, easy to implement from scratch, and flexible to modify and extend.
In this section, we present a means for embedding constraints into a spatio-temporal representation built on which the original routing problem can be tackled by efficient methods (though it is still a NP-hard problem).

As the problem essentially aims to determine the visiting time and order of vertices, it inspires us to incorporate a time dimension to extend the 2-dimensional graph on a spatial plane to a 3-dimensional graph (topology). To make the model implementable, the time range $[0, T]$ is discretized into a sequence of time intervals $\Delta_t$ of equal length, and the interval $\Delta_t$ is used as a time unit that specifies discrete time resolution. Thus, the travel time $\tau_{ij}$ can be expressed as multiple times of the time unit. For example, $\tau_{ij}=n_{ij}\Delta_t$ and $T=n_T\Delta_t$, where $n_{ij}$ and $n_T$  are integers.

Figure~\ref{fig:time_model} shows two examples that describe the basic idea.
Intuitively, one can imagine that each spatial graph vertex (Fig.~\ref{fig:n3t4_2}) is extended to a sequence of vertices expanding in the time dimension, where each vertex is on an unique time layer corresponding to a future time moment (Fig.~\ref{fig:n3t4}).
Then edges are added by concatenating vertices of different time layers, constrained by the real travel time.
Formally, two vertices on different time layers form an edge if and only if the two vertices are spatially traversable and the time difference between the two time layers is exactly equal to the anticipated real travel time between the two vertices.

Each vertex contains a profit and the profit is time-varying. In other words, the profits at different time layers are not the same. 
Therefore, the problem is equivalent to finding a path from the given start vertex at time $0$ such that the path transits other vertices within time $T$ and the total collected profits are maximized. 
It is worth mentioning that, the spatio-tempral edges are ``directed" since each edge must start from a vertex at an earlier time layer and ends at one at a later moment;
Because the time is uni-directional and cannot travel backwards, it is impossible to form a loop or cycle on the spatio-temporal graph. 
As a result, such a spatio-temporal representation is equivalent to a vertex-weighted {\em Directed Acyclic Graph (DAG)}~\cite{mahmoudi2016finding}. 
This allows us to conveniently develop our own routing algorithm built on many existing efficient DAG algorithms. 

More formally, the spatio-temporal representation is a vertex-weighted DAG, denoted by $G'=(V', E', W')$, where $V'=V\times\mathcal{T}$, $\mathcal{T}=\{0,\Delta_t,...,n_T\Delta_t\}$; $e_{ijus}$ is a directed edge from 
vertex $v'_{iu}=(v_i, u)$ to $v'_{js}=(v_j, s)$, and $e_{ijus}\in G'$ if and only if the edge $e(v_i, v_j)\in E$ in $G$ and $\tau_{ij} = n_s\Delta_t - n_u\Delta_t$; the weight $w'_{iu}\in W'$ for vertex $v'_{iu}$ has value $w'_{iu}=f_i(u)$. Keep in mind that $u=n_u\Delta_t$ and a similar equation holds for $s$. 

As an example, suppose that we have a complete graph $G$ with 3 vertices, $\Delta_t=1$, and $T=5$. In the spatio-temporal representation $G'$ shown in Fig.~\ref{fig:n3t4}, there are $18$ vertices $v'_{0,0}, v'_{0,1},...,v'_{2,5}$, each vertex in the original graph $G$ is duplicated at the discrete time in $\mathcal{T}=\{0,1,2,3,4,5\}$.
If $\tau_{01}=3$ and $\tau_{21}=5$, then vertex $v'_{0,0}$ is directed to $v'_{1,3}$ and $v'_{2,0}$ is directed to $v'_{1,5}$. In this example, we have $9$ directed edges in total.

With graph $G=(V,E)$ and given time limit $T$, the spatio-temporal graph $G'=(V',E', W')$  can be constructed by Alg.~\ref{algo:dag}. 
Assume there are $n$ vertices in $G$, the time complexity of  Alg.~\ref{algo:dag} is $O(n^2T/\Delta_t)$ because there are three for loops each of which is associated with either $n$ or $T$. 
Also there are $n(T/\Delta_t+1)$ vertices in $G'$, so the space complexity is $O(nT/\Delta_t)$ .





{
\small
\begin{algorithm}
\caption{SpatioTemporalGraph $(G, T, v_{0}, \Delta_t)$}
\label{algo:dag}
\begin{algorithmic}[1]
{\small
\REQUIRE
    \STATE Graph $G = (V, E)$, time limit $T$, start vertex $v_{0}$, time interval $\Delta_t$
\ENSURE
\FOR {each vertex $v_i$ in graph $G$}
    \FOR {each $t$ in time period $[0,T]$, time step is $\Delta_t$}
        \STATE $v'_{it}.id = i$
        \STATE $v'_{it}.weight = f_i(t)$, put $w'_{it}$ into the weight set $W'$
        \STATE $v'_{it}.position = v_i.position$
        \STATE $v'_{it}.profit = f_i(t)$
        \STATE $v'_{it}.sum = -\infty$
        \STATE $v'_{it}.parent = -1$
        \FOR{each $v_i$'s neighbor $v_j$}
            \IF {$t+ \tau_{ij}$ is equal to $t'$ AND $t'<=T$ }
                \STATE put edge $e_{ijtt'}$ into the edge set $E'$ 
                \STATE push $v'_{jt'}$ into vector $v'_{it'}.successors$ 
            \ENDIF
            \STATE update indegree of $v'_{it}$.
        \ENDFOR
        \STATE put $v'_{it}$ into the vertex set $V'$
    \ENDFOR
\ENDFOR
\STATE set the $v'_{0,0}.sum$ to 0
\STATE return graph $G' = (V', E', W')$
}
\end{algorithmic}
{
\vspace{-2 mm}
\line(1,0){100} \\
{\scriptsize
Note: $f_i(t)$ is the time-varying profit function of vertex $v_i$ at time $t$. 
$v'_{0,0}$ is the start vertex in $G'$, and $.sum$ is used to store accumulated profit from prior traversal along a path. The label $.parent$ points to the predecessor vertex.
}
}
\end{algorithm} 
}


\section{Routing Algorithm}

An important advantage of the spatio-temporal graph lies in that, 
the time related constraints have been incorporated into this spatio-temporal representation, so that 
temporal related constraints can be decoupled and eliminated from the process of optimizing the profits.
Thus, the problem~\eqref{eqt:f5}-\eqref{eqt:f11} is equivalent to finding a path $\mathcal{P}$ from vertex $v'_{0,0}=(v_0, 0)$ to $v'_{ks}$ on $G'$  where $s\leq T$ such that $\sum_{(v'_iu)\in \mathcal{P}}w'_{iu}$ is maximum.

Since the spatio-temporal graph is essentially a DAG, we develop our routing solution via extending classic DAG algorithms.
Specifically, we found that the profit maximization can be transformed to a {\em longest path} problem by accumulating the profits collected from vertices instead of summing up edge lengths along a path.
We manipulate the DAG so that vertices are sorted in a topological order along the temporal dimension, 
and then employ a {\em dynamic programming} paradigm to compute the maximal profit path.

It is also noteworthy that, while developing a solution to the time-varying OP,
we take two concerns that are related to applications into account.

\begin{itemize}
    \item \textit{Specification of a Routing Destination:} Many routing problems, including classic OPs, require to specify a routing destination. The destination can be the original depot where the agent departures (e.g., a mail truck needs to return a central processing office); the destination can also be an arbitrary depot located somewhere else (e.g., a freight truck needs to pick up goods and unload them into some specified processing location that is different from the starting depot). 
    \item \textit{Persistent Task:} Many long-term missions need repetitive and persistent routing, for which specifying a routing destination is not necessary or even inappropriate. 
    For instance, in the persistent environmental monitoring task, we do not need the robot to stop at some specified location, as the robot will need to resume to next round of routing after the completion of current one. 
    Thus, the routing destination should be computed on the fly based on the profit optimization constrained by time $T$, instead of being manually specified.
\end{itemize}

We will show that the proposed framework works for both specified and unspecified routing destinations.


\subsection{Topological Sorting of DAG in Temporal Dimension}

The main purpose of topological sorting of the DAG is that, the vertices are ``placed" onto different ``stages" according to their temporal constraints, so that a dynamic programming structure (discussed in the following subsection) can be applied.

Formally, a topological sort of a directed graph is a linear ordering of its vertices such that for every directed edge $e_{uv}$ from vertex $u$ to vertex $v$, $u$ comes before $v$ in ordering.
A topological sort of a graph requires that the graph must be a DAG. 
We employ a well-known algorithm developed by Kahn~\cite{Kahn1962} to sort our spatial-temporal graph $G'$, with main steps shown in
Alg.~\ref{algo:topological_sort}.
Briefly, 
we first find a list of vertices that have no incoming edges (with $deg^-(v)=0$), and insert them into a set $S$. Note, at least one such vertex must exist in a non-empty graph.
Then we traverse the set $S$. Each time we remove a vertex $v$ from $S$, and add it to the tail of the list $L$. After removing $v$, the indegree of its successors should be decreased by 1. Then we insert those vertices with updated indegree equal to 0 in the set $S$.

To analyze the time complexity, assume there are $n$ vertices in $G$, so there will be $n(T/\Delta_t+1)$ vertices in $G'$. 
For each vertex $v'_{it}$, there will be at most $n-1$ directed edges. 
Therefore $|E'| = O(n^2T/\Delta_t)$, and the time complexity of Alg.~\ref{algo:topological_sort} is $O(|V'|+|E'|)=O(n^2T/\Delta_t)$.

{
\small
\begin{algorithm}
\caption{TopologicalSort ($G'$) }
\label{algo:topological_sort}
\begin{algorithmic}[1]
{\small
\STATE $L$: an empty list that will contain the sorted elements
\STATE $S$:  a set of all nodes with no incoming edges 
\WHILE {$S$ is not empty}
	\STATE remove a vertex $v$ from $S$
	\STATE add $v$ to the end of $L$
	\FOR {each successor $u$ of $v$}
	    \STATE $deg^-(u)=deg^-(u)-1$ 
	    \IF { $deg^-(u)$ equals to 0}
		    \STATE insert $u$ into $S$
	    \ENDIF
	\ENDFOR
\ENDWHILE
\STATE return $L$ (a topologically sorted order)
}
\end{algorithmic}
{\vspace{-4 mm}
\line(1,0){100} \\
{\scriptsize
{\bf Note:} $G'$ is a DAG, the indegree of $v$ is denoted as $deg^-(v)$.}
}
\end{algorithm} 
}


\subsection{Computing Maximal Profit Path}

We transfer the time-varying OP to a longest path problem in a DAG. 
The classic longest path problem is the problem of finding a simple path of maximum length in a given graph.
We employ a dynamic programming structure to memorize incumbent maximal accumulated profit at each vertex of topologically sorted stages.
Note that, in our problem, we need to optimize profits collected from vertices, instead of adding up length of edges. 
Therefore, instead of using the longest path update function between two successive stages:
\begin{equation}
l[w]=l[v]+\tau_{vw}, \text{ if } l[v]+\tau_{vw}>l[w], 
\end{equation}
where $l[w]$ is the largest distance from start vertex to $w$, 
we utilize an update function: 
\begin{equation}
w.sum = v.sum+w.profit, \text{ if } v.sum+w.profit>w.sum.
\end{equation}

The computation of the maximal profit path is described in Alg.~\ref{algo:dp}.
Briefly, 
the topologically sorted vertices $V'$ of the spatio-temporal graph $G'$ are used as an input.
Then vertices from different stages form dynamic programming subproblems and they are updated with accumulated profits recursively, starting from $v'_{0,0}$. 
Here we use $v'_{it}.path$ to store vertices along the path from start $v'_0$ to $v'_i$ (Note, only spatial information is recorded). 
To prevent from forming routing subtours, we check and discard those already visited vertices before each value update.

After the completion of dynamic programming, each vertex contains information of the maximal profit path that routes from $v'_{0,0}$ to it. 
Since the graph $G'$ has incorporated the time limit $T$, every vertex is feasible to the time constraint.
To find a maximal profit path to a specified vertex (destination) $v_i \in G$, one simply needs to enumerate all states $v'_{ij} \in G', \forall{j}$ of vertex $v_i$ and retrieve the path with the largest value. 
If a destination is not specified, one needs to enumerate all states of all vertices and find out the maximal one among them.
The time complexity of Alg.~\ref{algo:dp} is $O(n^2T/\Delta_t)$ due to its two for loops.


{
\small
\begin{algorithm}
\caption{$MaximalProfitPath(L)$ }
\label{algo:dp}
\begin{algorithmic}[1]
{\small
 \STATE $v_{it}.sum=0, i=0, t=0$
 \FOR {each vertex $v'_{it}$ in topologically sorted order $L$}
    \FOR {each vertex $v'_{jt'}$ in $v'_{it}.successors$}
        \IF {vertex $v_j$ is not in the $v'_{it}.path$}
            \IF {$v'_{it}.sum+v'_{jt'}.profit>v'_{jt'}.sum$}
                \STATE $v'_{jt'}.sum = v'_{it}.sum+v'_{jt'}.profit$
                \STATE update the $v'_{jt'}.parent$ to $v'_{it}$
                \STATE $v'_{jt'}.path = v'_{it}.path + v_j$ 
            \ENDIF
        \ENDIF
    \ENDFOR
 \ENDFOR
 \FOR{each vertex $v'_{it}$ in $G'$}
    \STATE find the maximal $sum$
 \ENDFOR
 \STATE retrieve a path $P$ by backtracking from end to start
 \STATE return $P$
}
\end{algorithmic}
{\vspace{-2 mm}
\line(1,0){100} \\
{\scriptsize
{\bf
Note:} The label $.path$ is a vector that stores all visited vertices from start to $v_i$\\
}
}
\end{algorithm} 
}


\subsection{Main Routing Algorithm}

With the components descried above, the main algorithm is shown in Alg.~\ref{algo:main}. According to the analysis of each part, the overall time complexity of our algorithm is $n^2T/\Delta_t$, and the space complexity is $n^2T/\Delta_t$ as well.

{
\small
\begin{algorithm}
\caption{Time-Varying OP}
\label{algo:main}
\begin{algorithmic}[1]
{\small
    \REQUIRE 
      \STATE 2D graph $G=(V,E)$, start vertex $v_0$, time limit $T$, time interval $\Delta_t$
    \ENSURE
    \STATE construct a spatio-temporal graph \\ $G'=SpatioTemporalGraph(G, T, v_{0}, \Delta_t)$
    \STATE $L=TopologicalSort(G')$
    \STATE $path = MaximalProfitPath(L)$
    \STATE return $path$
}
\end{algorithmic}
\end{algorithm} 
}

Our algorithm has a polynomial time complexity given a fixed $\Delta_t$, but cannot guarantee to find the optimal solution to this NP-hard problem.
The main reason for possibly reaching at sub-optimality lies in that,
we added a non-subtour constraint in order to (1) prevent the path from traversing back and forth among a small subset of adjacent vertices and (2) comply with the structure of dynamic programming.
Such a constraint eliminates certain feasible searching space which possibly contains the optimal solution.
One may regard this as a trade-off between solution quality and practical runtime.
Nevertheless, our evaluation results with extensive trials show that on average our algorithm produces near-optimal solutions.

\subsection{Discussion: Effects of Discretization}
The time interval $\Delta_t$ plays a critical role in our framework as the total profits are actually evaluated at travel time steps equal to multiple $\Delta_t$. Yet it may be impractical to select a small enough time interval because it would result in an extremely large size of DAG. One possible implementation is to round the travel time to a value closet to some integer times of $\Delta_t$. Hence, it is important to analyze the impact of errors due to such rounding procedure. 

Assume that the profit function $f_i(t)$ satisfies the following Lipschitz condition:
\begin{equation}\label{eqt:Lip} 
|f_i(t) - f_i(s)| \leq K |t-s|, ~\forall t, s\in[0, T], \forall i\in\{0,...,n\},
\end{equation}
where $K$ is a constant independent of $i$. Suppose that the optimal objective value for the problem \eqref{eqt:f5}-\eqref{eqt:f11} is $z$, and the optimal objective value by the proposed framework is $z'$. Then we have the result as follows. 

\begin{proposition}
Under the condition \eqref{eqt:Lip}, we have the upper bound for the difference in two objective values:  
\begin{equation}\label{eqt:errorBound} 
|z-z'|\leq [n(n+1)/2]\cdot K\cdot \Delta_t.
\end{equation}
\end{proposition} 
\textit{proof:} Let $\mathcal{P}$ be the optimal path in the graph $G$ for the problem \eqref{eqt:f5}-\eqref{eqt:f11} and let $\mathcal{P'}$ be the optimal path in the proposed spatio-temporal DAG $G'$. By the construction of $G'$, $P$ must correspond to a path $\tilde{P}$ in the graph $G'$. Let $\tilde{z}$ be the total profits (i.e. the value of objective~\eqref{eqt:f5} on this path) for the path $\tilde{P}$. Since the difference between the discrete time in our framework and the original real-valued time is within $\Delta_t$, the fact that any path contains no more than $n$ vertices and the condition \eqref{eqt:Lip} lead to $|z-\tilde{z}|\leq [n(n+1)/2]K\Delta_t$. As $P'$ rather than $\tilde{P}$ is the optimal path in $G'$, we have $\tilde{z}\leq z'$. By combining these results, we should have $z-z'\leq [n(n+1)/2]K\Delta_t$.
Conversely, by construction any feasible path on the graph $G'$ must correspond to a feasible path on $G$. Using similar arguments, we have $z'-z\leq [n(n+1)/2]K\Delta_t$. Therefore, the result \eqref{eqt:errorBound} holds.  
It is worthwhile noting that the constraint that the path in $G$ or $G'$ contains no subtours is critical to the result \eqref{eqt:errorBound}.


\section{Results}

We conducted numerical evaluations to validate the proposed algorithm.
We compared our algorithm with other algorithms, all of which were implemented in C++.
The experiments were performed on a system with an Intel i5 2.2GHz processor and 4GB RAM.
In our experiments, vertices are randomly generated in the area $[-50, 50]\times[-50, 50]$ on the $x$-$y$ plane,
and each vertex has a profit (weight) following certain time-varying functions.
The distance between two vertices is Euclidean distance.


\subsection{Routing Feature of ``Orienteering"}

\begin{figure} \vspace{-10pt}
  \centering
 \subfigure[]
 	{\label{fig:dis}\includegraphics[height=1.5in]{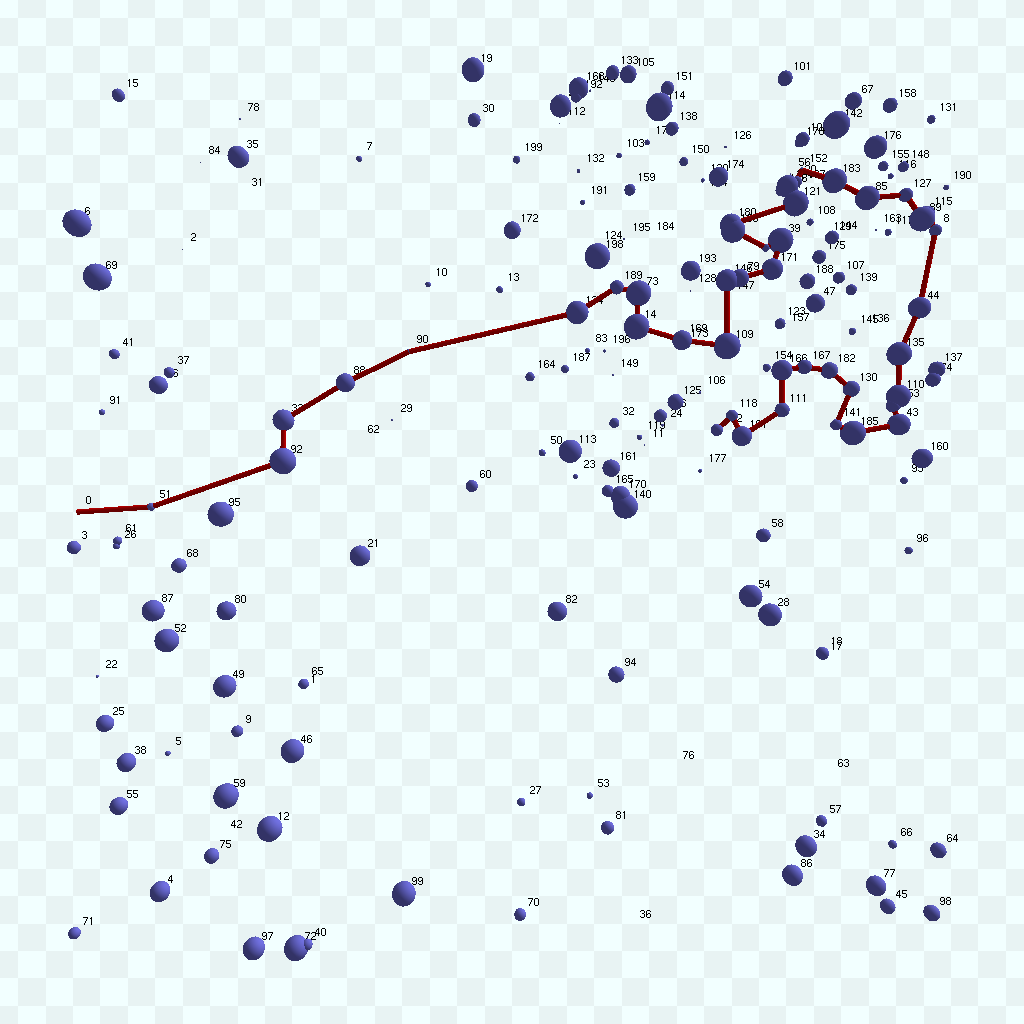}} 
 	\quad
  \subfigure[]
  	{\label{fig:step_func}\includegraphics[width=1.5in]{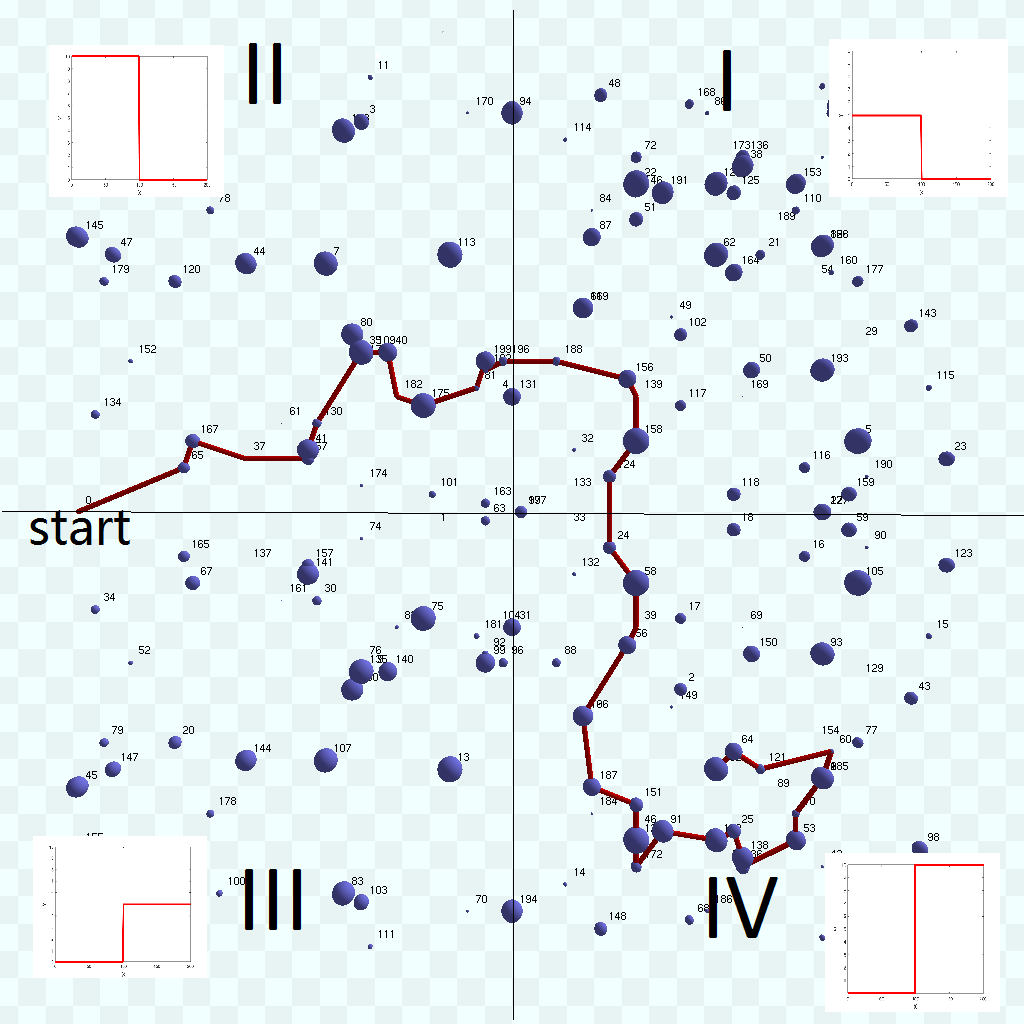}} \vspace{-8pt}
  \caption{(a) Route feature of orienteering. $\Delta t$ is 1, vertex number is 200, $T$ is 200 and the start vertex is at (-49,0); (b) Route behavior under time-varying profit function, $\Delta t$ is 1, vertex number is 200, $T$ is 200. The start vertex $v_0$ is at (-49,0). In all of our experimental graphs, the size of a vertex represents the weight, which is proportional to its profit.}
\label{fig:route-ft}
\end{figure}

First, we demonstrate that our solution has the basic orienteering property: given limited time, the path always goes through those vertices with large profits.
Fig.~\ref{fig:dis} shows an example.
We purposely manipulate the profit distribution among the vertices, such that in the upper right corner, there are more vertices with larger profits.
For better visualization, in the figure the size of a vertex is proportional to the profit on it. 
Additionally, we purposely put the starting vertex on the left and do not specify a destination. 
From Fig.~\ref{fig:dis} we can see that, in order to obtain more profits, the route goes to the upper-right corner almost directly from the start vertex. 
In this way, a big part of time is used among traversing those vertices with large profits. 

\subsection{Routing Properties under Temporal Variations}

To examine the effects caused by temporal variations, we manipulate the profit functions.
Specifically, 
we divide the space into four regions (quadrants) I,II,III and IV, as shown in Fig.~\ref{fig:step_func}.
The profits are the same if the vertices are in the same region, but different if not.
For example, in region I, the profit function for each $v'_{it}$ is 
\[ f_i(t)_I =
  \begin{cases}
    5w_i       & \quad \text{if } t \leqslant T/2 \\
    0  & \quad \text{if } t > T/2\\
  \end{cases}
\]
where $w_i$ is the weight of vertex $v_i$ in graph $G$. (Keep in mind that, $w_i$ is different from $w'_{it}$ which is time-varying. Here the weight $w_i$ refers to a fixed parameter for profit function.)
Similarly, the profit functions in regions II, III and IV are 
\[ f_i(t)_{II} =
  \begin{cases}
    10w_i       & \quad \text{if } t \leqslant T/2 \\
    0  & \quad \text{if } t > T/2\\
  \end{cases}
\]

\[ f_i(t)_{III} =
  \begin{cases}
    0       & \quad \text{if } t \leqslant T/2 \\
    5w_i  & \quad \text{if } t > T/2\\ 
  \end{cases}
\]

\[ f_i(t)_{IV} =
  \begin{cases}
    0       & \quad \text{if } t \leqslant T/2 \\
    10w_i  & \quad \text{if } t > T/2\\
  \end{cases}
\]

Figure~\ref{fig:step_func} reveals that, the path first transits the vertices in region II, because during the first half $T$, vertices in region II have larger profits than those in region III. 
After the path enters region I and after the time passes $T/2$, vertices in region IV contain larger profits.
Such variations attract the path to go through vertices in region IV.
This example indicates that our algorithm is sensitive to the time-varying profit functions.


\subsection{Comparison with Optimal Solution}

\begin{figure*} \vspace{-10pt}
  \centering
 \subfigure[$f(t) = w_it/T$]
 	{\label{fig:linear_func}\includegraphics[height=1.2in]{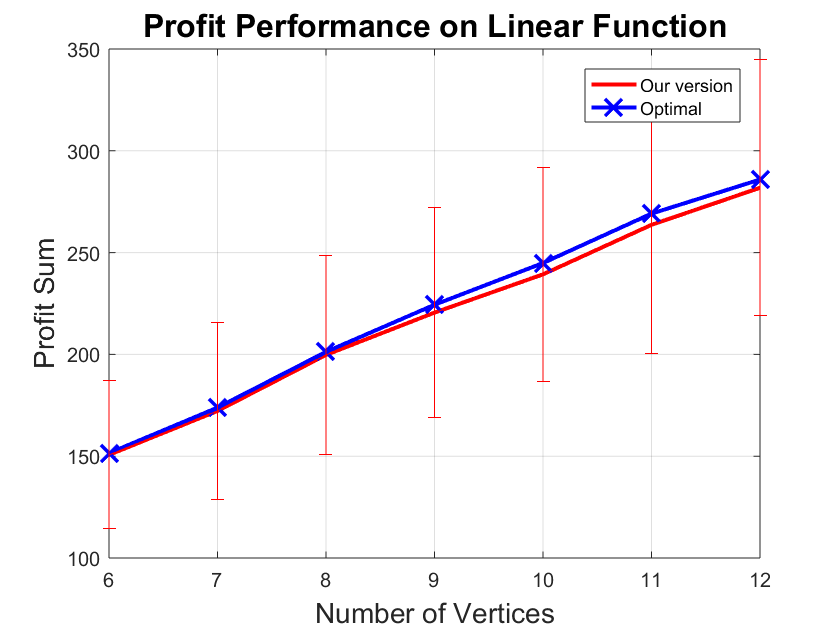}} 
 	\quad \quad
  \subfigure[$f(t) = -w_i(t^2+tT+T^2)/T^2 $]
 	{\label{fig:concave_func}\includegraphics[height=1.2in]{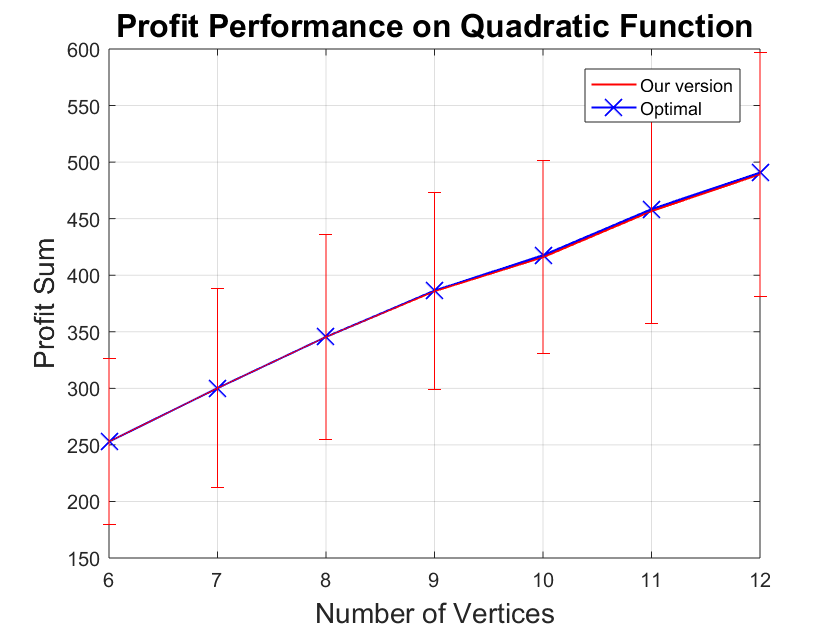}}
 	\quad \quad
  \subfigure[$f(t) = w_ilog(t+1) $]
  	{\label{fig:log_func}\includegraphics[height=1.2in]{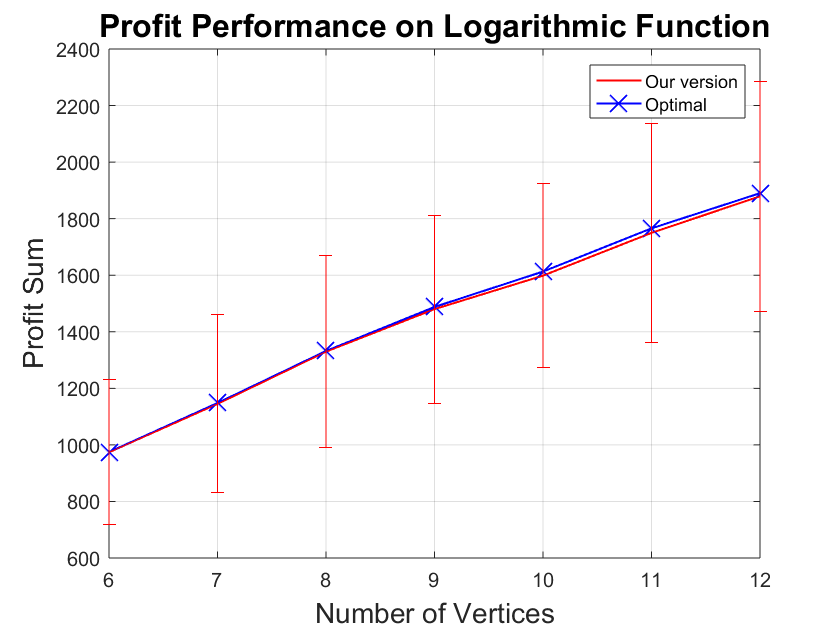}} \vspace{-8pt}
  \caption{Profit statistics on (a) Linear function; (b) Quadratic function; (c) Logarithmic function. }
\label{fig:implementation2} \vspace{-10pt}
\end{figure*}

We compare the result of our algorithm with the optimal solution that is obtained by enumerating all solutions in a brute-force way. 
Because of the prohibitive time complexity $O(n!)$ for searching for the optimal solution, the practical runtime for 13 vertices requires more than 10 minutes.
Thus, we tested up to 12 vertices to compare with the optimal solutions. 
We investigated three representative profit functions in the form of linear, quadratic and logarithmic, respectively. 
The results are shown in Fig.~\ref{fig:implementation2}. We can see that the results of our algorithm are very close to those of the optimal solutions, for all the three functions. 

\begin{figure*}[t] 
  \centering
  \subfigure[]
  	{\label{fig:search5}\includegraphics[height=1in]{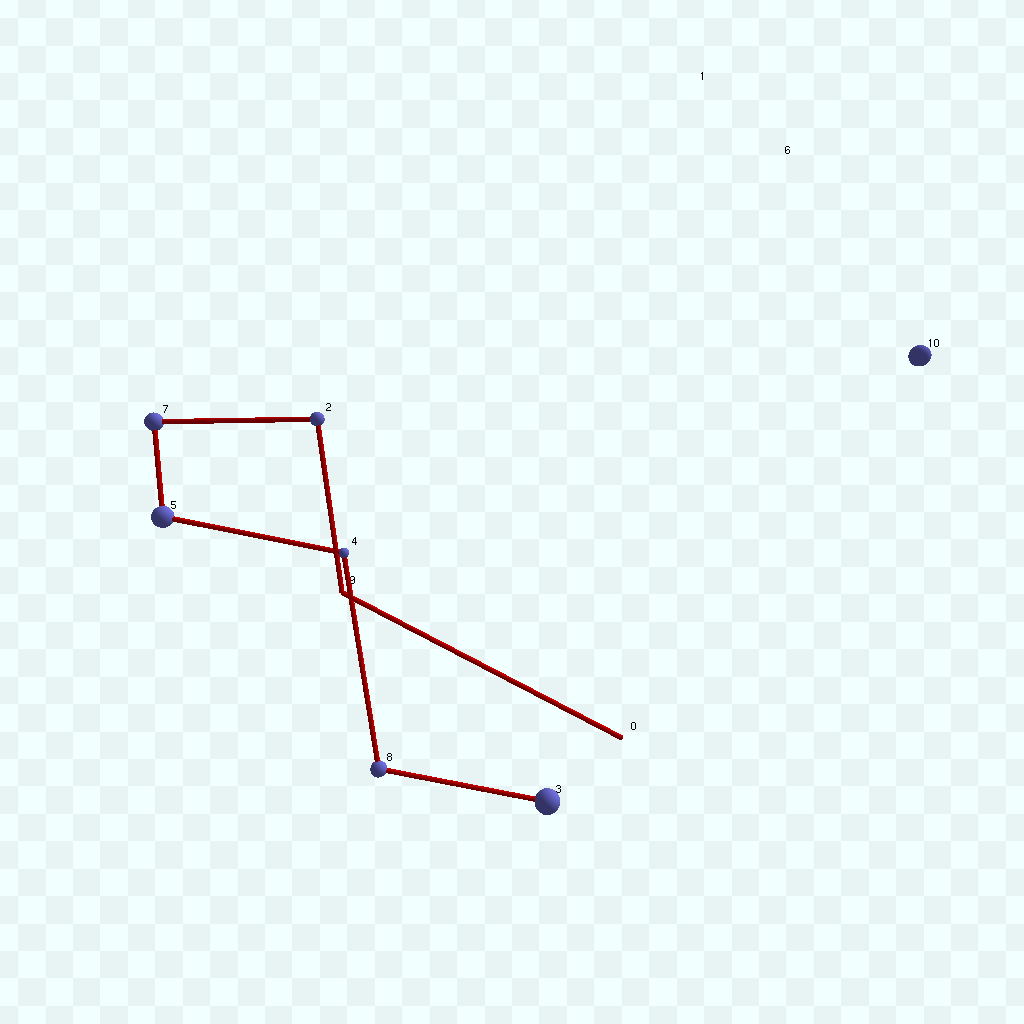}} 
  \subfigure[]
 	{\label{fig:search6}\includegraphics[height=1in]{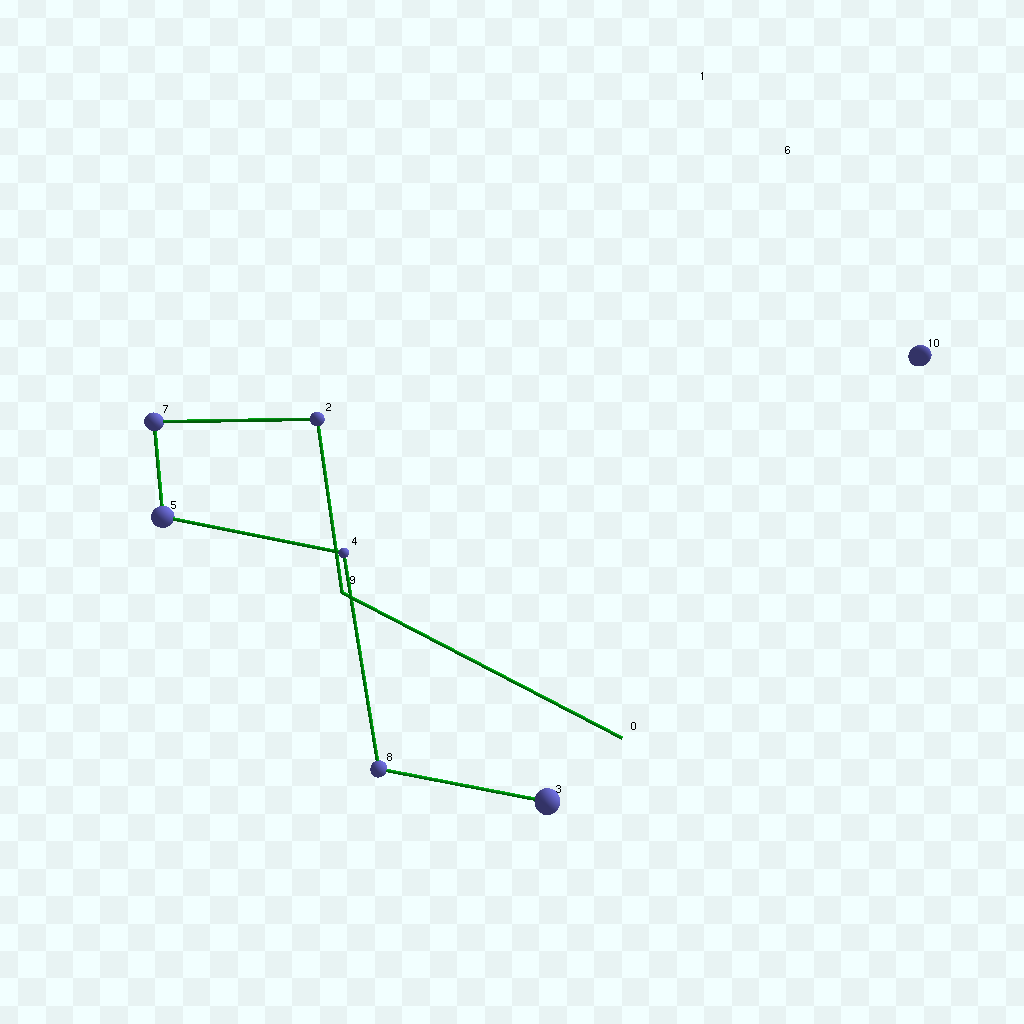}}
 \subfigure[]
 	{\label{fig:search1}\includegraphics[height=1in]{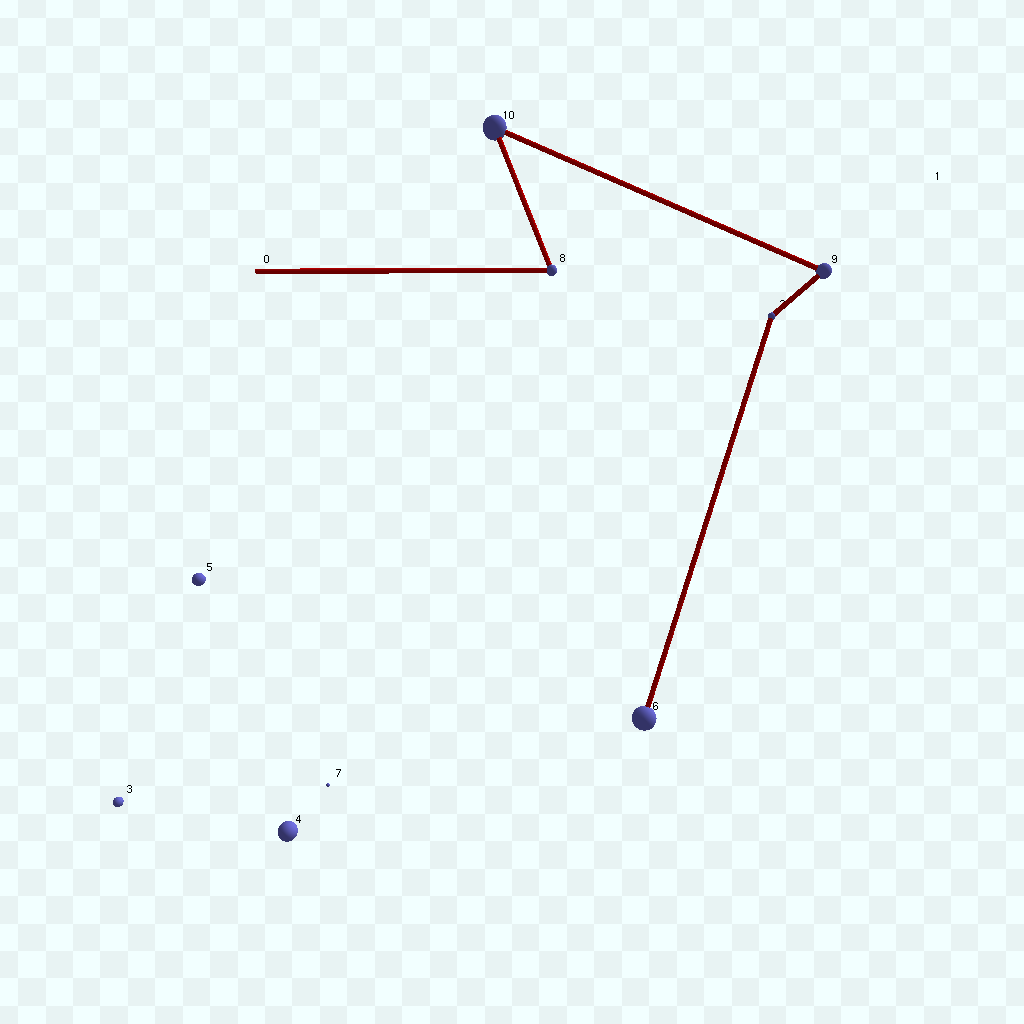}} 
  \subfigure[]
 	{\label{fig:search2}\includegraphics[height=1in]{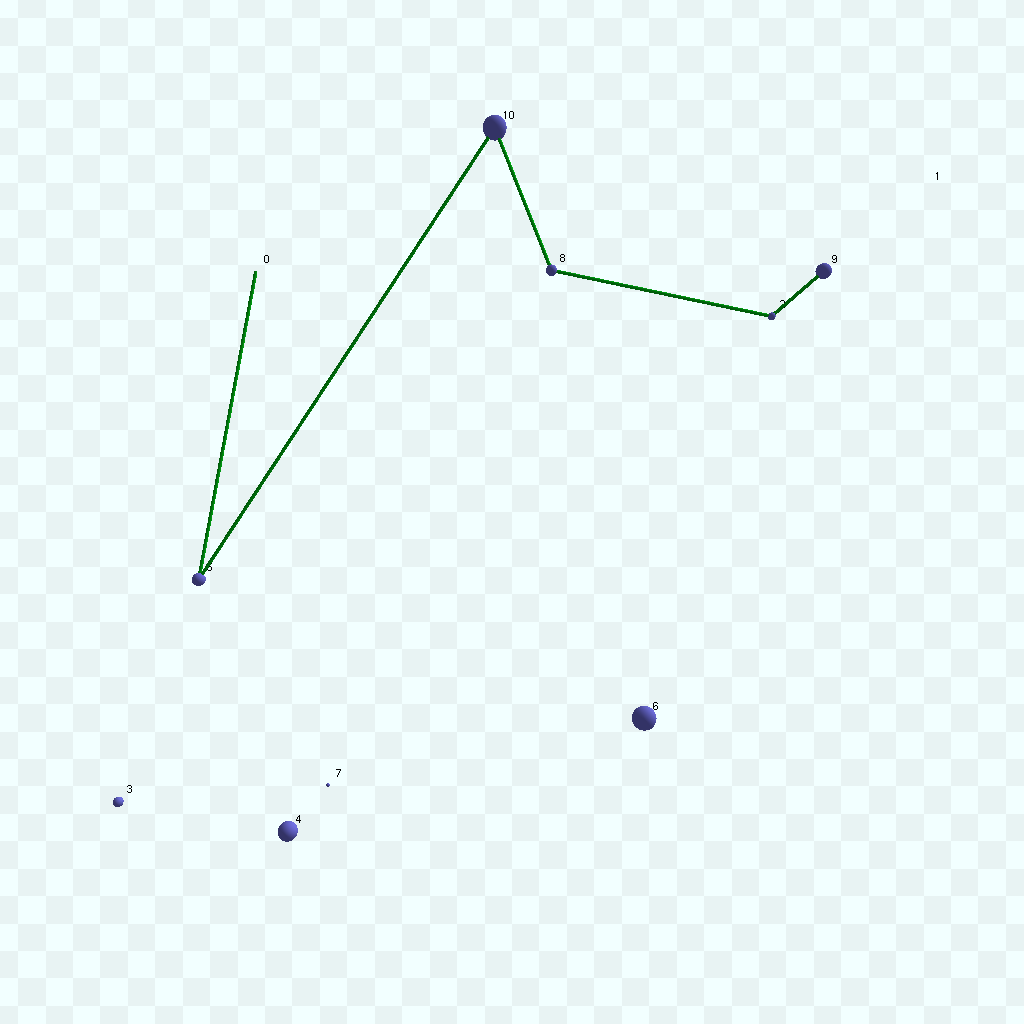}}
  \subfigure[]
  	{\label{fig:search3}\includegraphics[height=1in]{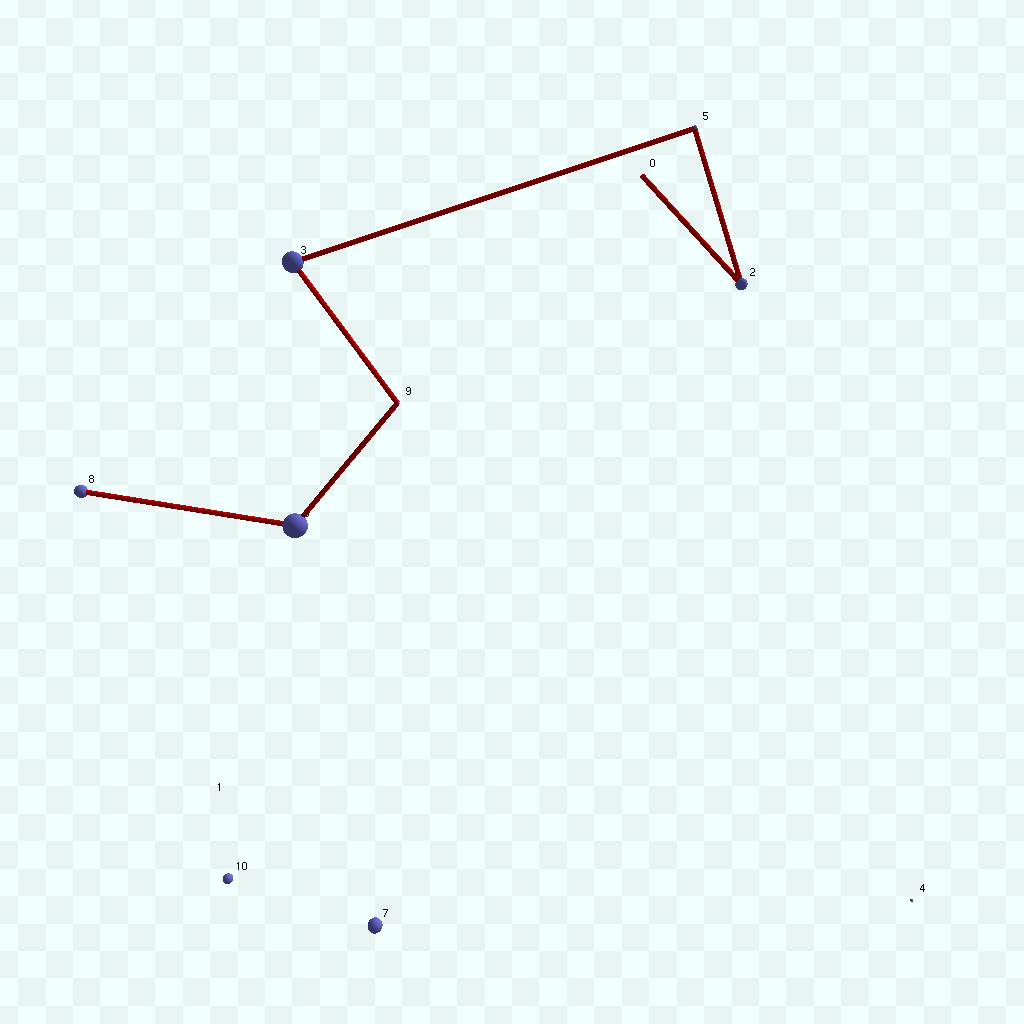}} 
  \subfigure[]
 	{\label{fig:search4}\includegraphics[height=1in]{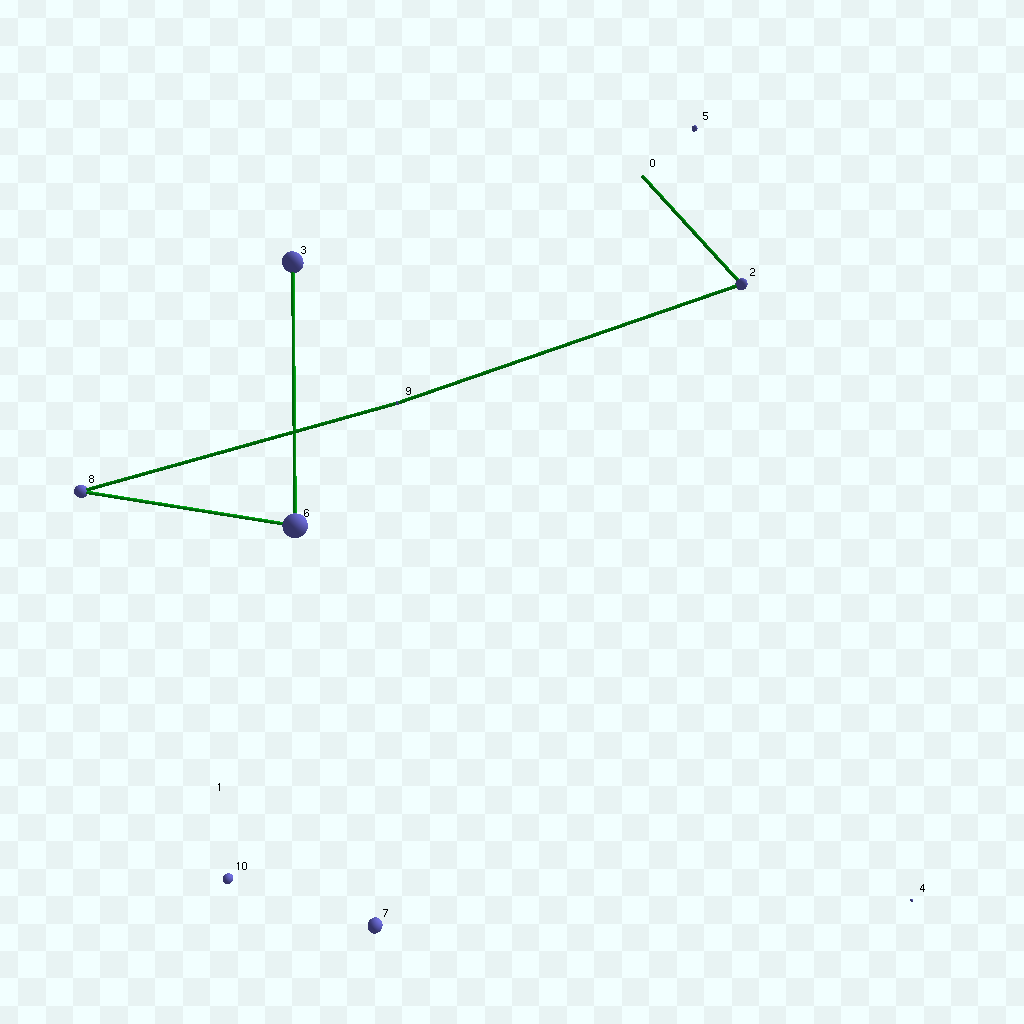}}\vspace{-8pt}
  \caption{Examples with linear profit functions. The paths in red are from our algorithm while the paths in green are found with brute-force search. (a)(b) Our algorithm obtains optimal solution 281; (c)(d) Our solution is 183.5 whereas the optimal solution is 185.0; (e)(f) Our solution is 186.6 whereas the optimal solution is 195.1. }
\label{fig:example_opt} \vspace{-10pt}
\end{figure*}



Figure~\ref{fig:example_opt} shows a group of paths produced from our algorithm and the optimal solution.
We use a linear profit $f_i(t)=w_it/T$ for this example. 
In many cases, our algorithm get the same results as those of the optimal solution, as shown in Fig.~\ref{fig:search5} and Fig.~\ref{fig:search6}.
Sometimes the paths are different, such as Fig.~\ref{fig:search1} and Fig.~\ref{fig:search2}, but the collected profits are quite similar (in this example, ours obtains 183.5 whereas the optimal is 185.0). Fig.~\ref{fig:search3} and Fig.~\ref{fig:search4} show another example.



\subsection{Comparison with Classic OP Algorithm}

\begin{figure*}
  \centering
 \subfigure[]
 	{\label{fig:our_static}\includegraphics[height=1.3in]{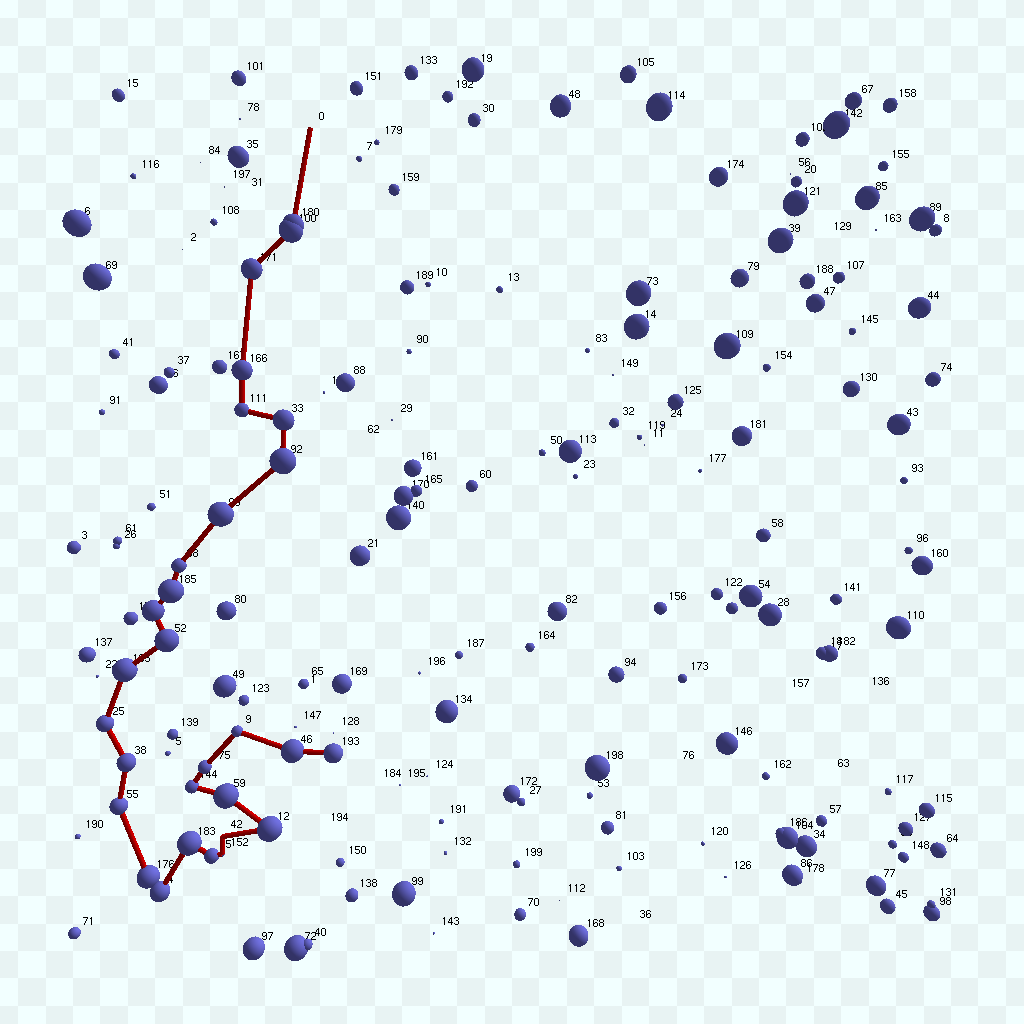}} 
 	\quad \quad
 \subfigure[]
  	{\label{fig:op_static}\includegraphics[height=1.3in]{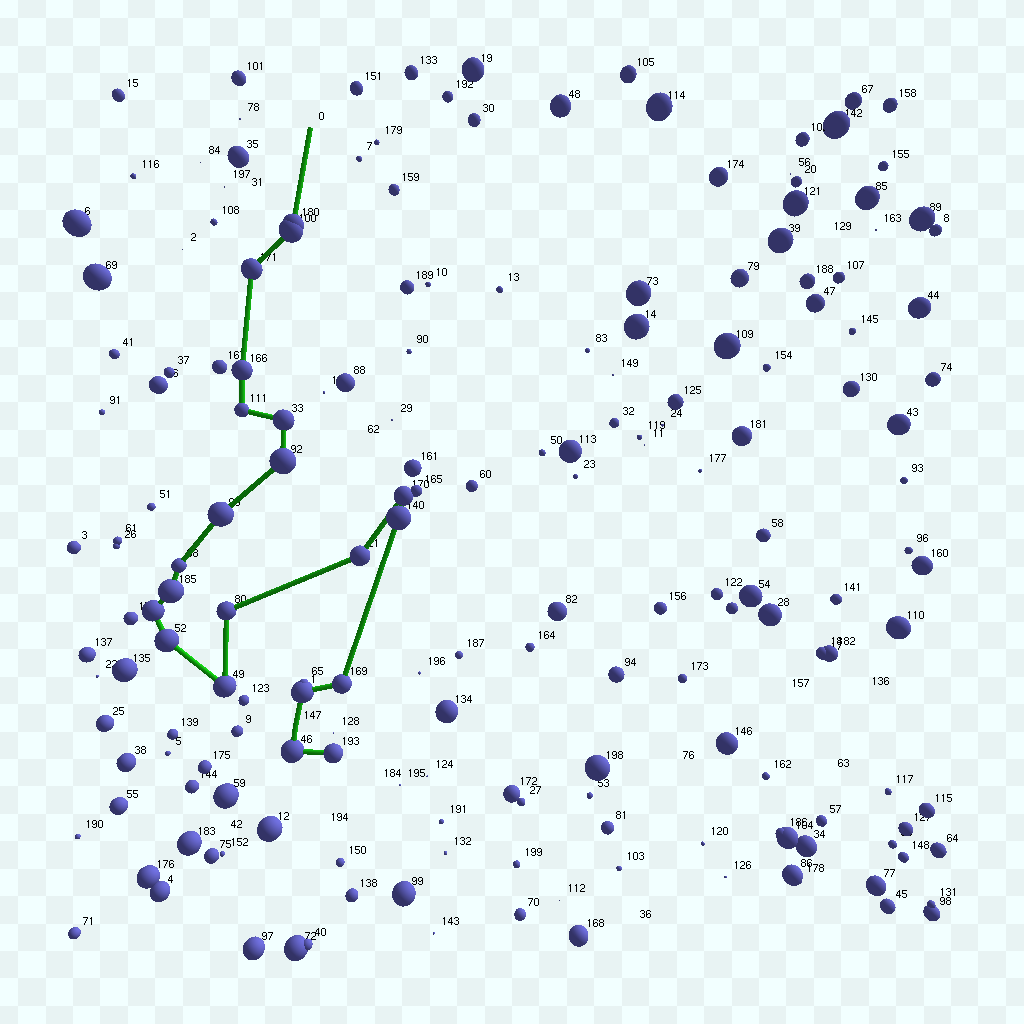}}
  	\quad \quad
  \subfigure[]
 	{\label{fig:our_linear}\includegraphics[height=1.3in]{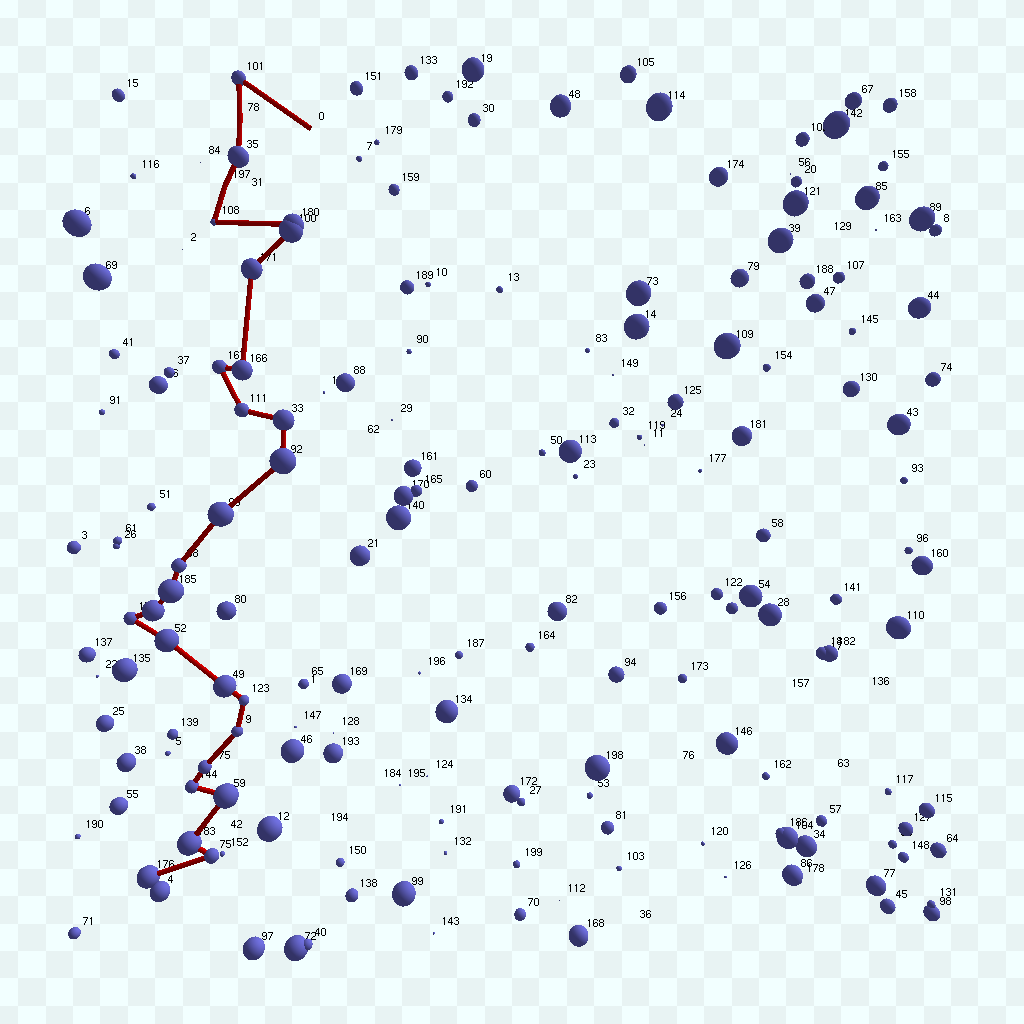}}
 	\quad \quad
  \subfigure[]
  	{\label{fig:op_linear}\includegraphics[height=1.3in]{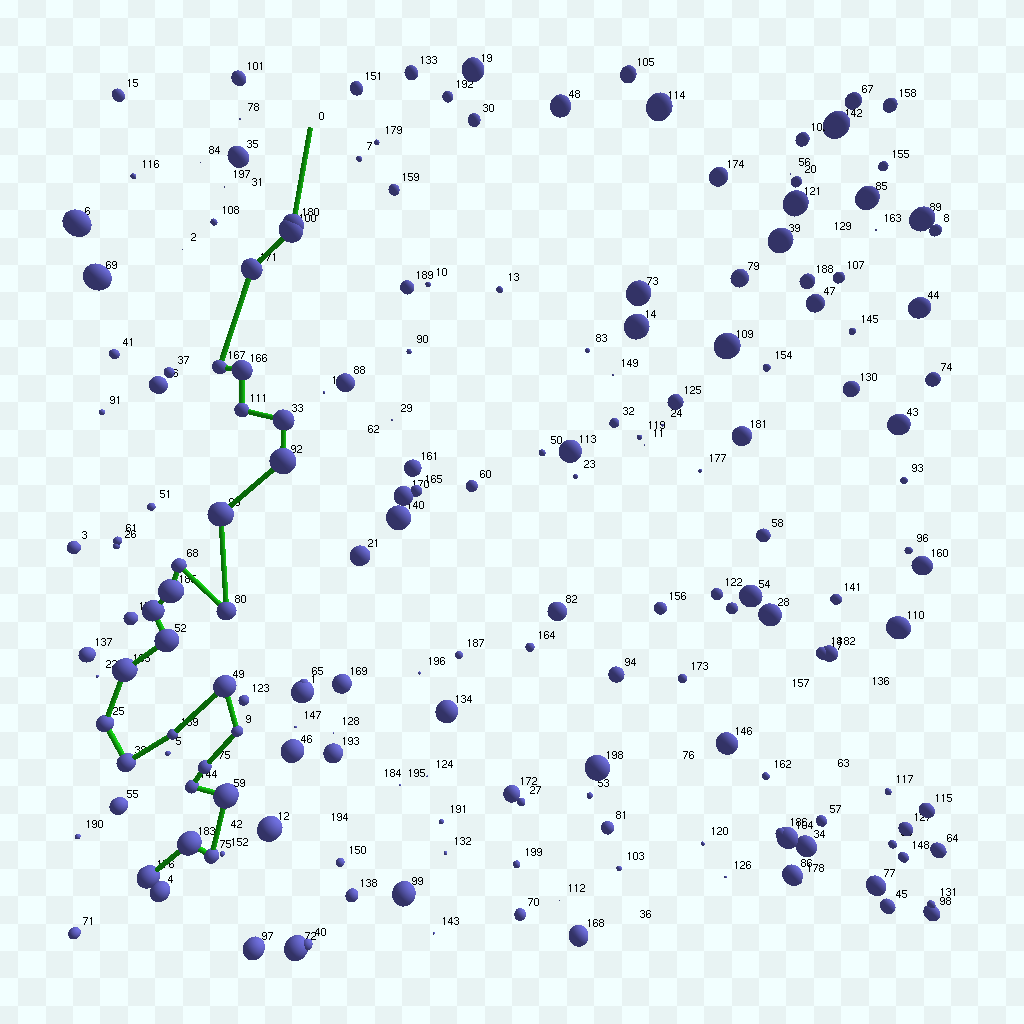}} \vspace{-8pt}
  \caption{Path comparisons. The paths in red are computed from our algorithm while the paths in green are from the classic OPs using the center-of-gravity heuristic. (a)(b) Time-invariant profits $f_i(t)=w_i$; (c)(d) Time-varying profits $f_i(t)=w_it/T$}
\label{fig:compare_op_demo} \vspace{-10pt}
\vspace{-10pt}
\end{figure*}

\begin{figure}
  \centering
 \subfigure[]
 	{\label{fig:time_invariant}\includegraphics[width=1.6in]{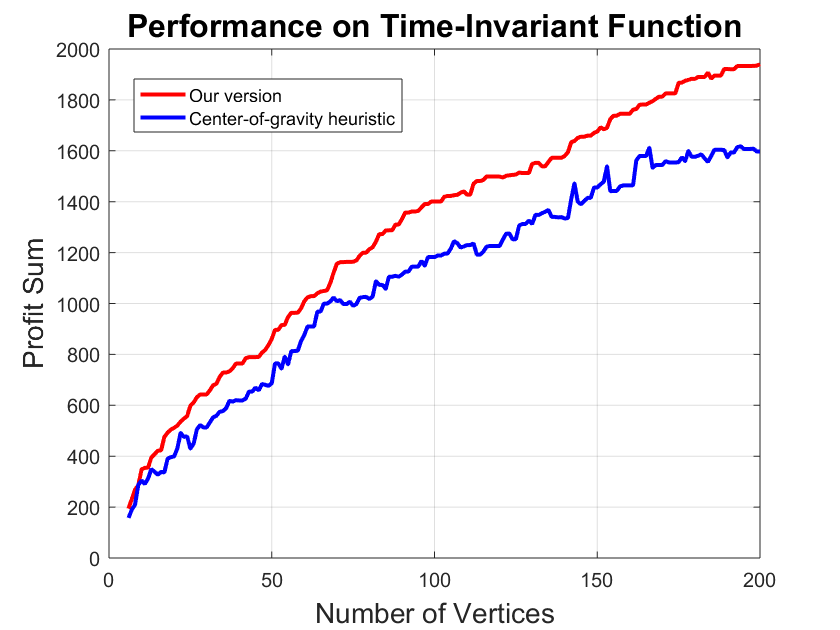}} 
 	\quad 
 \subfigure[]
  	{\label{fig:time_variant}\includegraphics[width=1.6in]{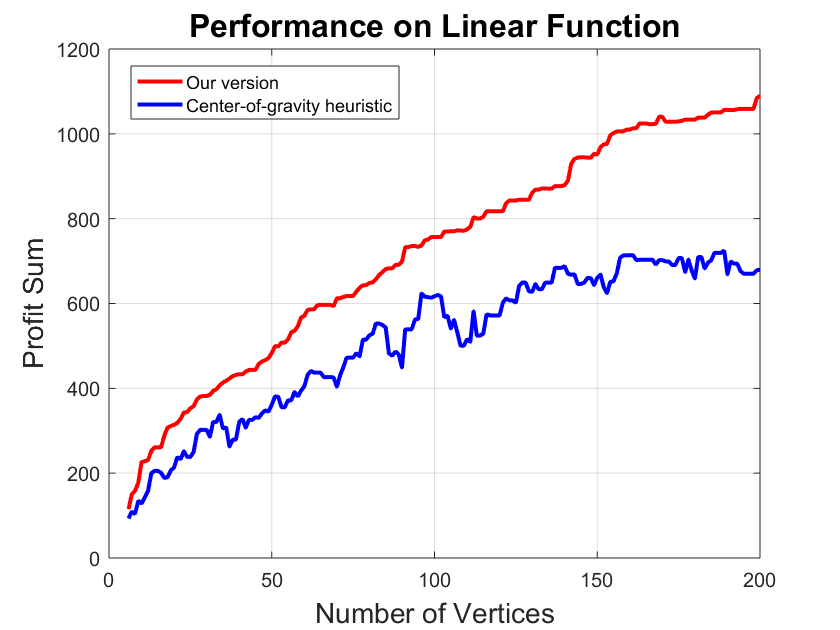}}\vspace{-10pt}
 \subfigure[]
 {\label{fig:running_time2} \includegraphics[width=1.6in]{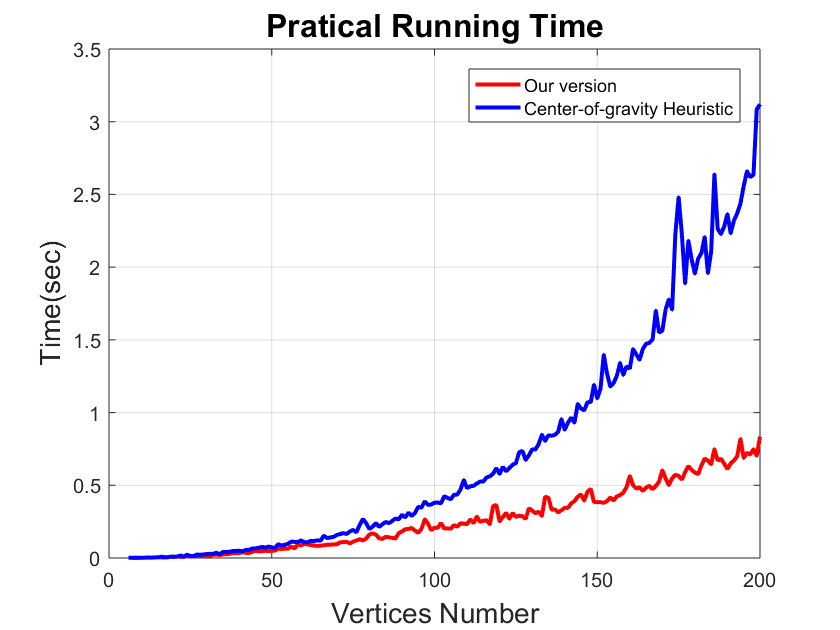}} \vspace{-8pt}
  \caption{
  (a) Performance on time-invariant profits $f_i(t)=w_i$; (b) Performance on time-varying profits $f_i(t)=w_it/T$; (c) Statistics of practical running time. }
\label{fig:compare_op} \vspace{-10pt}
\end{figure}

We also compared our algorithm with the classic OP algorithms. 
We implemented a well-known heuristic called {\em center-of-gravity}~\cite{golden1987orienteering}, and tested with both time-invariant and time-variant profits. 
Fig.~\ref{fig:compare_op_demo} demonstrates the differences between the results of our approach and those of the OP.
Fig.~\ref{fig:compare_op} show statistics of numerical evaluations with 200 vertices. 
We can see that, our algorithm outperforms the OP for both the time-invariant and time-varying profits.
Particularly, the difference margin is larger for the time-varying profits. 
Actually, the classic OP cannot handle time-varying profits and the total profit curve quickly converges (with fixed given time $T$, the overall profit of OP stops growing along with the increment of the number of vertices).

The running time is compared in Fig.~\ref{fig:running_time2}, from which we can observe that, in practice our algorithm costs less time than that of the OP using the center-of-gravity heuristic.

\subsection{Path Quality under Different Time Intervals}

Time interval $\Delta_t$ determines time discretization resolution, and therefore affacts the optimality as well. 
Table~\ref{tab:table1} shows statistics of the route's quality under different time intervals. 
In the table, numbers in the first row are interval values, and numbers in the first column are the number of vertices. The remaining values are the collected profits from our algorithm.
We can see that, in general the route's quality is better when the time interval is smaller. 
However, obviously a smaller time interval will inevitably lead to a larger graph and thus require a larger running time.

\begin{table}[h]
  \centering
  \caption{Statistics of solution quality with various time intervals $\Delta_t$}
  \vspace{-3pt}
  \label{tab:table1}
  \renewcommand{\arraystretch}{1.5}
  \begin{tabular}{>{\centering\bfseries}m{0.35in} >{\centering}m{.29in} >{\centering}m{.29in} >{\centering}m{.29in} >{\centering}m{.29in} >{\centering}m{.29in} >{\centering\arraybackslash}m{.29in}}
    \toprule
    Intervals & \textbf{0.1} & \textbf{0.5} & \textbf{1} & \textbf{2} & \textbf{5} & \textbf{10} \\
    \midrule
    50	&541.1	&530.3	&529.2	&522.1	&515.1 &498.3 \\
    100	&788.2	&783.4	&767.9	&735.6	&727.5	&619.2 \\
    150	&950.2	&919.0	&909.7	&892.9	&823.0	&686.8 \\
    200	&1129.5	&1117.6	&1107.1	&1094.3	&935.5	&696.0 \\
    \bottomrule
  \end{tabular}\\
  {\small 
    \vspace{2pt}
    Note: $T=150$, graph within $[-50, 50]\times[-50, 50]$ on $x$-$y$ plane, profit $f_i(t)=w_it/T.$ 50,100,150 and 200 are the numbers of vertices.
  }
  \vspace{-2mm}
\end{table}

\subsection{Demonstration with New York Taxi Data}

\begin{figure}\vspace{-10pt}
  \centering
   	\includegraphics[height=1.6in]{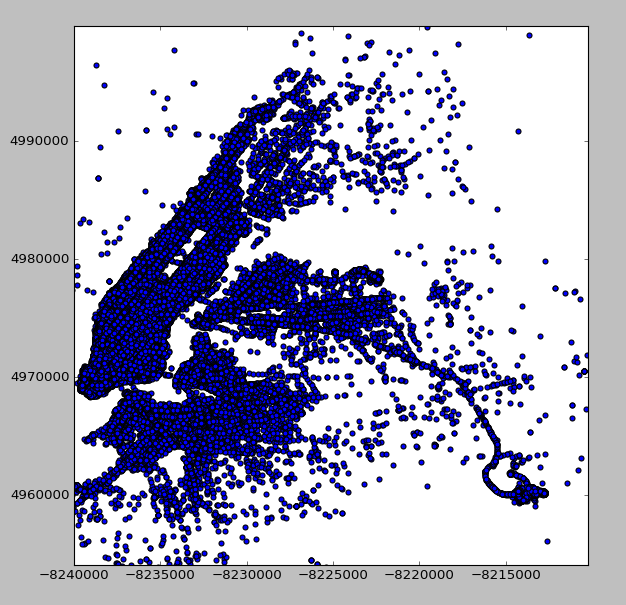}  
   \caption{Data of taxi calls in New York city}
\label{fig:ny1}
\end{figure}

\begin{figure}
 \vspace{-5mm}
  \centering
 \subfigure[]
  	{\label{fig:ny2}\includegraphics[height=1.1in]{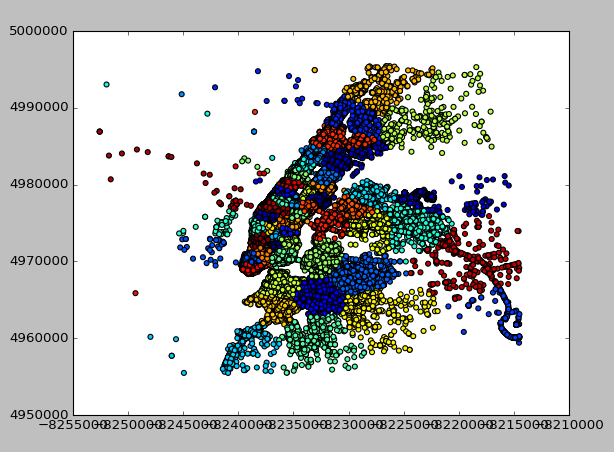}} 
 \subfigure[]
  	{\label{fig:ny3}\includegraphics[height=1.1in]{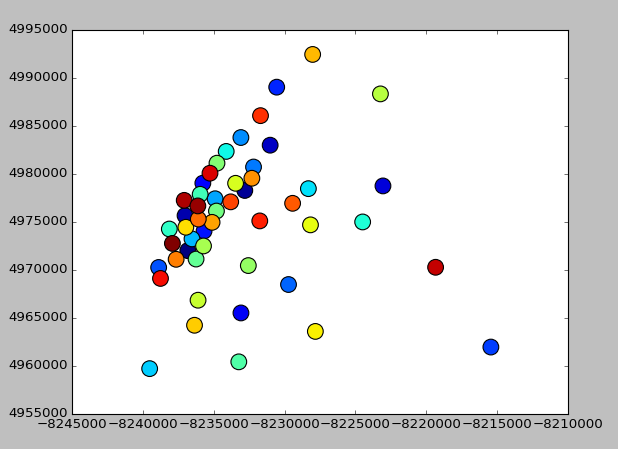}} \vspace{-8pt}
  \caption{
  (a) Taxi calls are clustered through $k$-means; (b) Clustering centers represent high-level routing depots. }
\label{fig:ny_op}
\end{figure}

\begin{figure}
  \centering
 \subfigure[]
 	{\label{fig:taxi0}\includegraphics[width=1.5in]{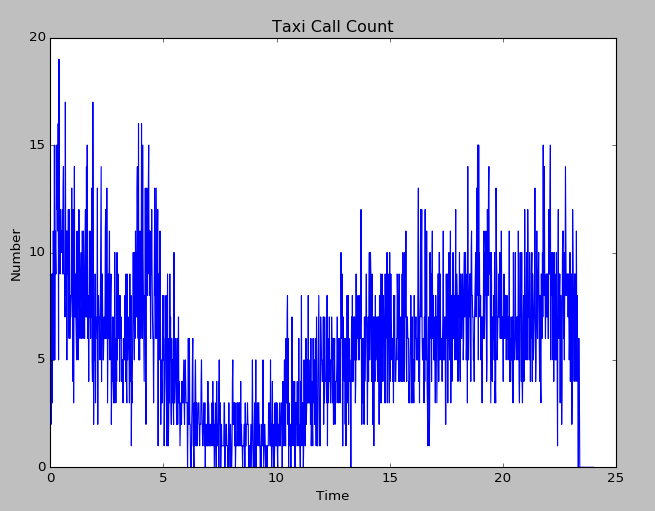}} 
 \subfigure[]
  	{\label{fig:taxi1}\includegraphics[width=1.5in]{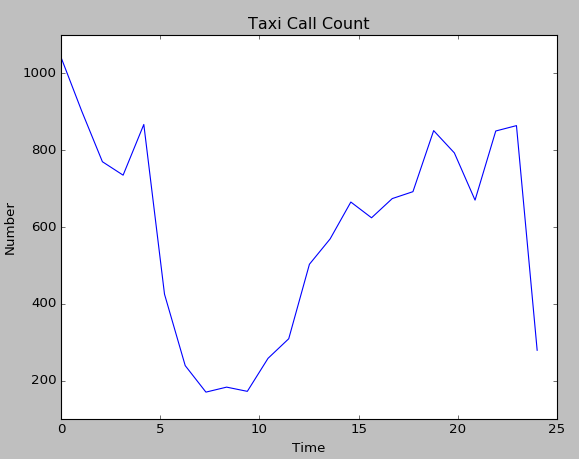}} \vspace{-8pt}
  \caption{
  For a particular region (region 1 in this example), the number of taxi calls is varying as time elapses. (a) Time unit is minute; (b) Time unit is hour.}
\label{fig:taxi_call} 
\end{figure}

\begin{figure} [t] 
  \centering
   	\includegraphics[height=2in]{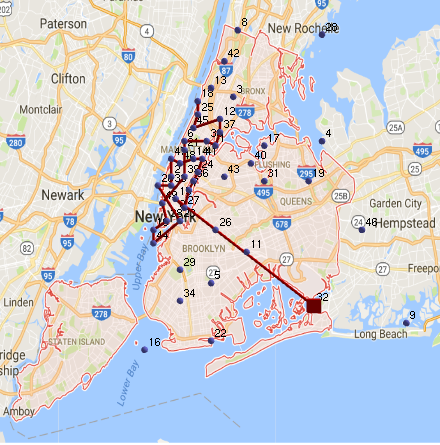}  \vspace{-8pt}
   \caption{The taxi routing path generated from our algorithm. The big square is the starting location.}
\label{fig:ny4} 
\end{figure}

We demonstrate that the proposed algorithm can be used for (autonomous) taxi routing.
Fig.~\ref{fig:ny1} shows real data of taxi calls in New York city. The data is obtained from NYC Taxi and Limousine Commission~\cite{NYC-taxi}.
We use the $k$-means clustering algorithm to cluster the whole New York taxi calls into 50 local regions, as shown in Fig.~\ref{fig:ny2}. The clustering centers are shown in Fig.~\ref{fig:ny3}.
We assume that each pick-up and the subsequent drop-off occur in the same local region, and each pick-up and drop-off is counted as a completion of one service.  
Then a taxi driver aims at maximizing the profit by providing more services during a fixed period.
In each region, we calculate the total number of taxi calls in every predefined time unit (e.g., minute). An example is shown in Fig.~\ref{fig:taxi_call}. 
We regard the total number of taxi calls in a region as a time-varying profit, and more calls indicate higher chance of getting  profits (note, there are other taxis too).  
Therefore, this problem can be formulated as an orienteering problem with time-varying profits, where we need to treat those clustered regions as super routing depots that offer different profits at different times of a day. 
Fig.~\ref{fig:ny4} demonstrates a result generated from running our algorithm, from which we can observe that the taxi is routed across those regions with the most taxi calls, even though we set the starting point at a distant location with sparse requests. 

\section{Conclusion and Future Work}

We presented a framework for addressing the orienteering problem with time-varying profits.
Instead of following traditional mixed integer program solution routines, 
we develop an intuitive and effective framework that incorporates time-variations into the fundamental planning process. 
Specifically, we first construct a deterministic {\em spatio-temporal} representation where both spatial description and temporal logic are unified into one routing topology, and then we extend existing sorting and searching algorithms, so that the routing solutions can be computed in an extremely efficient way. 
Finally, we validated our algorithm with numerical evaluations and the results show that our framework produces near-optimal solutions in a very efficient way.


{ 
\bibliographystyle{abbrv}
\bibliography{reference.bib}
}

\end{document}